\documentclass[10pt,twocolumn,letterpaper]{article}

\usepackage{cvpr}              

\usepackage{float}
\usepackage{multirow}
\usepackage{makecell}
\usepackage{adjustbox}

\usepackage{xcolor}
\usepackage{fvextra}
\usepackage{fancyvrb}
\usepackage{marvosym}
\usepackage[most]{tcolorbox}

\definecolor{redfg}{HTML}{D35B27}
\definecolor{redbg}{HTML}{FFF7F3}
\definecolor{yellowfg}{HTML}{FABB00}
\definecolor{yellowbg}{HTML}{FFF7E1}
\definecolor{bluefg}{HTML}{0C69DA}
\definecolor{bluebg}{HTML}{FBFCFE}

\definecolor{cvprblue}{rgb}{0.21,0.49,0.74}
\usepackage[pagebackref,breaklinks,colorlinks,allcolors=cvprblue]{hyperref}

\def\paperID{8601} 
\def\confName{CVPR}
\def\confYear{2025}

\title{ComfyBench: Benchmarking LLM-based Agents in ComfyUI for Autonomously Designing Collaborative AI Systems}

\author{
Xiangyuan Xue$^{1,2}$\textsuperscript{*} \quad
Zeyu Lu$^{1,2}$ \quad
Di Huang$^{3}$ \quad
Zidong Wang$^{1,4}$ \quad
Wanli Ouyang$^{1,4}$ \quad
Lei Bai$^{1}$\textsuperscript{\Letter} \\
$^{1}$Shanghai Artificial Intelligence Laboratory \quad $^{2}$Shanghai Jiao Tong University \\
$^{3}$The University of Sydney \quad $^{4}$ The Chinese University of Hong Kong
}

\begin{document}

\maketitle

\renewcommand{\thefootnote}{}
\footnotetext[1]{\textsuperscript{*}This work was done during his internship at Shanghai Artificial Intelligence Laboratory.\quad \textsuperscript{\Letter}Corresponding author, baisanshi@gmail.com.}

\begin{abstract}
Much previous AI research has focused on developing monolithic models to maximize their intelligence, with the primary goal of enhancing performance on specific tasks. In contrast, this work attempts to study using LLM-based agents to design collaborative AI systems autonomously. To explore this problem, we first introduce ComfyBench to evaluate agents's ability to design collaborative AI systems in ComfyUI. ComfyBench is a comprehensive benchmark comprising 200 diverse tasks covering various instruction-following generation challenges, along with detailed annotations for 3,205 nodes and 20 workflows. Based on ComfyBench, we further develop ComfyAgent, a novel framework that empowers LLM-based agents to autonomously design collaborative AI systems by generating workflows. ComfyAgent is based on two core concepts. First, it represents workflows with code, which can be reversibly converted into workflows and executed as collaborative systems by the interpreter. Second, it constructs a multi-agent system that cooperates to learn from existing workflows and generate new workflows for a given task. While experimental results demonstrate that ComfyAgent achieves a comparable resolve rate to o1-preview and significantly surpasses other agents on ComfyBench, ComfyAgent has resolved only 15\% of creative tasks. LLM-based agents still have a long way to go in autonomously designing collaborative AI systems. Progress with ComfyBench is paving the way for more intelligent and autonomous collaborative AI systems. Our code is available at: \href{https://github.com/xxyQwQ/ComfyBench}{https://github.com/xxyQwQ/ComfyBench}.
\end{abstract}    
\section{Introduction}
\label{section:introduction}

\begin{figure*}
    \centering
    \includegraphics[width=1\linewidth]{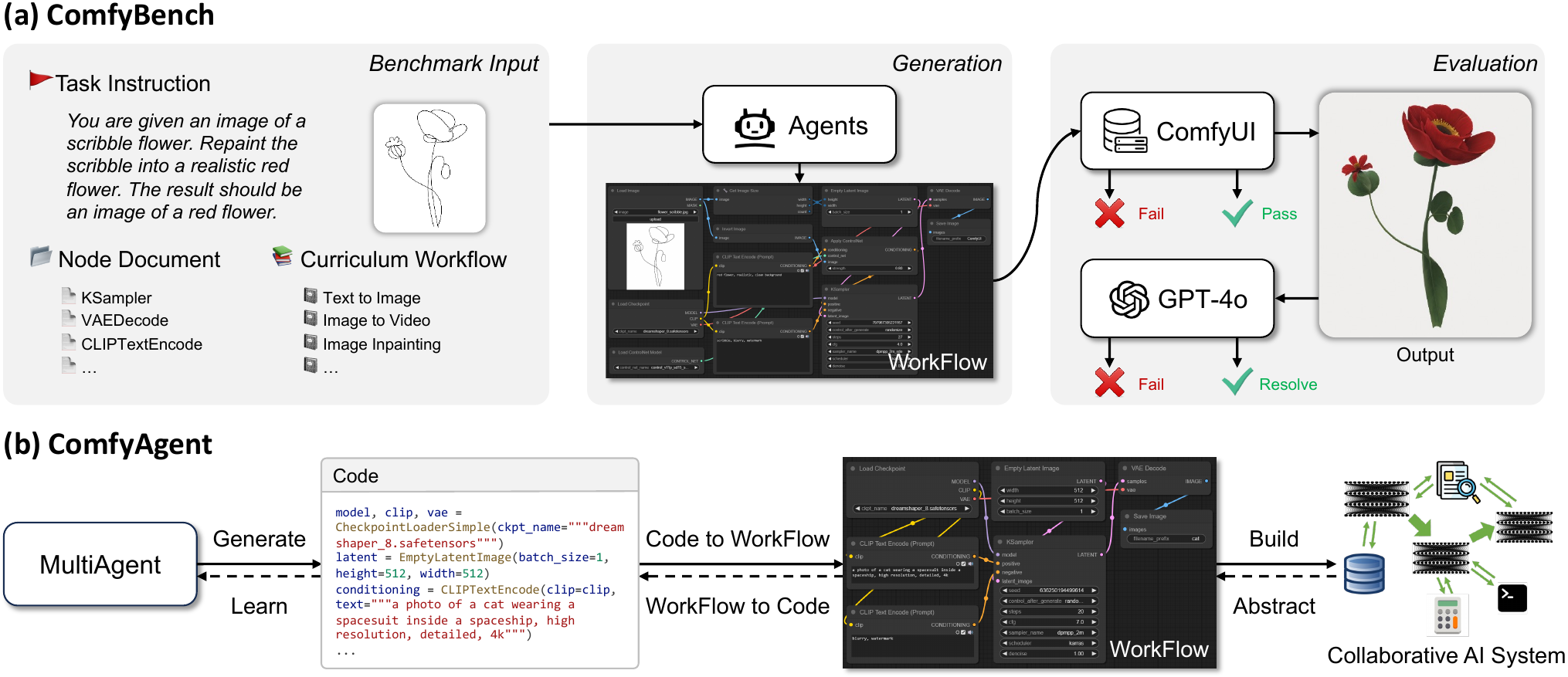}
    \caption{(a) ComfyBench is a comprehensive benchmark to evaluate agents's ability to design collaborative AI systems in ComfyUI. Given the task instruction, agents are required to learn from documents and create workflows to describe collaborative AI systems. The performance is measured by \textit{pass rate} and \textit{resolve rate}, reflecting whether the workflow can be correctly executed and whether the task requirements are realized. (b) ComfyAgent builds collaborative Al systems in ComfyUI by generating workflows. The workflows are converted into equivalent code so that LLMs can better understand them. ComfyAgent can learn from existing workflows and autonomously design new ones. The generated workflows can be interpreted as collaborative AI systems to complete given tasks.}
    \label{figure:teaser}
\end{figure*}

The recent evolution of AI is defined by the growing importance of collaborative AI systems, which integrate multiple models and tools to work as a whole collaborative system~\citep{matei2024shift}. The success of ChatGPT~\citep{openai2024chatgpt} presents the possibility of integrating a wide range of tasks, such as web browsing~\citep{nakano2021webgpt}, image generation~\citep{ramesh2021zero}, and code execution, into a single conversational agent. Unlike traditional AI models that function as single entities, collaborative AI systems integrate multiple AI components, each contributing unique capabilities to solve complex problems. This shift towards integration has become crucial for achieving state-of-the-art results, as it leverages the combined strengths of diverse AI functionalities within a unified framework.

Unfortunately, the design space and optimization of collaborative systems often require significant human expertise~\citep{li2024autoflow}. For instance, AlphaCode 2~\citep{AlphaCode2T} utilizes an intricate AI system to achieve expert-level performance in competitive programming, which involves fine-tuned LLMs for coding and scoring, expert models for clustering, and modules for code execution and filtering. These components collaborate in a precise manner carefully designed by human experts to achieve state-of-the-art performance, which is not achievable by any single component or their casual combination. This brings us to a pivotal question: \textit{Can we develop an agent, akin to a human expert, capable of autonomously designing collaborative AI systems?}

In this work, we explore this question in ComfyUI~\citep{comfyanonymous2023comfyui}, a trending and open-source platform that supports various generative tools and models in terms of nodes. Within ComfyUI, users can flexibly construct and connect nodes to build the workflow of the collaborative AI system, which is capable of producing highly customized artwork. 

To evaluate agents's ability to design collaborative AI systems in ComfyUI, we introduce \textbf{ComfyBench}, a comprehensive benchmark shown in Figure~\ref{figure:teaser}, which comprises 200 diverse tasks covering various instruction-following generation challenges, along with detailed annotations for 3,205 nodes and 20 workflows. We categorize tasks into three levels of difficulty: \textit{vanilla}, \textit{complex}, and \textit{creative}. In each task, agents are required to create workflows that describe collaborative AI systems within ComfyUI, which can then be executed to produce the desired outcomes specified in the task instructions. We assess the performance of agents using two evaluation metrics: \textit{pass rate}, showing the success of workflow execution, and \textit{resovle rate}, reflecting the effectiveness in meeting task requirements.

Based on ComfyBench, we find such tasks present great challenges for agents. First, the native JSON representation of workflows is inadequate for LLM-based agents to fully comprehend the intrinsic logic and dependencies within the systems~\citep{xu2024core}. Second, properly organizing the pipeline and utilizing AI-related modules require substantial expert knowledge and complex cognitive processes~\citep{hu2024automated}. Traditional methods, such as Chain-of-Thought (CoT)~\citep{wei2022chain} or Retrieval-Augmented Generation (RAG)~\citep{lewis2020retrieval}, suffer from context limitation and hallucination problems, thus failing to generate complex workflows reliably.

To tackle the aforementioned problems, we introduce \textbf{ComfyAgent}, a novel framework shown in Figure~\ref{figure:teaser}, which empowers LLM-based agents to autonomously design collaborative AI systems by generating workflows. The core concepts behind ComfyAgent are two-fold. First, it represents workflows with code, which can be reversibly converted into workflows and executed as collaborative systems by the interpreter. Second, it constructs a multi-agent system~\citep{guo2024large} which cooperates to learn from existing workflows and generate new workflows for a given task. The core of the multi-agent system is PlanAgent, responsible for the global planning of workflows by task instruction. After collecting document descriptions of all components and code examples of collected workflows, RetrievalAgent automatically retrieves and learns key information about the given task from code. CombineAgent and AdaptAgent are responsible for code modification to construct a workflow that can fulfill the specified task. Upon task completion, PlanAgent finishes the task to form the final workflow.

We evaluate common baseline agents~\citep{brown2020language,wei2022chain,wang2022self,lewis2020retrieval}, as well as our proposed ComfyAgent, on ComfyBench. The experimental results indicate that ComfyAgent achieves comparable \textit{pass rate} and \textit{resolve rate} to o1-preview~\citep{openai2024introducing} and significantly surpasses other agents on ComfyBench. However, ComfyAgent has resolved only $15\%$ of creative tasks, which is still far from satisfactory. We believe that LLM-based agents still have a long way to go for autonomously designing collaborative AI systems.

Our contributions can be summarized as follows:
\begin{itemize}
\item We introduce \textbf{ComfyBench}, the first-of-its-kind comprehensive benchmark for the development of agents capable of designing and executing a wide range of collaborative AI systems in ComfyUI.
\item We propose \textbf{ComfyAgent}, which represents workflows with code and constructs a novel multi-agent framework to empower LLM-based agents in autonomously designing collaborative AI systems.
\item Experimental results demonstrate that with our intricate multi-agent design, ComfyAgent achieves a comparable pass rate and resolve rate to o1-preview and significantly surpasses other agents on ComfyBench.
\end{itemize}
\section{Related Work}
\label{section:related_work}

\subsection{LLM-based Agents}
\label{section:llm_based_agents}

LLM-based agents leverage LLMs to interact with external tools and solve real-world problems~\citep{xi2023rise,wang2024survey}. Agents can exploit the powerful capabilities of LLMs in cognition and creativity, adapting to various novel tasks without relying on human intervention~\citep{li2023camel,mei2024llm}. Besides, agents have access to a wide range of external tools, making up for the natural limitation of processing only natural language~\citep{chen2024mindsearch,zhang2023data}. Researchers have been devoted to enhancing the capabilities of LLM-based agents. Many prompt-based methods~\citep{song2023llm,besta2024graph,yao2022react,wei2022chain,wang2022self,madaan2024self,wang2024executable} are proven effective for improving agents' planning and reasoning. Memory mechanism~\citep{zhong2024memorybank,zhang2024survey} and RAG~\citep{lewis2020retrieval,gao2023retrieval} further extend agents' working context and knowledge base. Reinforcement learning~\citep{ouyang2022training,bai2022training,kaufmann2023survey}, serving as a post-training paradigm, allowing even stronger enhancement and alignment. These various methods hasten the advent of advanced theories for agents, such as self-evolution~\citep{zhou2024symbolic,xu2024interactive,xi2024agentgym} and unified architecture~\citep{zhou2023agents,sumers2023cognitive}.

Based on well-designed module architectures and interaction mechanisms, LLM-based agents can be applied to solve a wide range of complex tasks, such as web navigation~\citep{nakano2021webgpt,yao2022webshop,zhou2023webarena,deng2024mind2web,koh2024visualwebarena}, interface operation~\citep{yang2023appagent,xie2023openagents,xie2024osworld,cao2024spider2,wu2024copilot}, code generation~\citep{wang2022execution,yin2022natural,yang2024swe,wang2024opendevin}, and other tasks in specified domains~\citep{huang2024blenderalchemy,thakur2023language,shen2024hugginggpt,ge2024openagi,yang2024intercode}. These tasks, previously considered to require expert knowledge and human intervention, can now be accomplished by agents reliably.

\subsection{Collaborative AI Systems}
\label{section:collaborative_ai_systems}

\citet{matei2024shift} points out that state-of-the-art results are increasingly obtained by collaborative AI systems instead of monolithic models, which has been verified by many previous works~\citep{AlphaCode2T,trinh2024solving,lewis2020retrieval,nori2023can,team2023gemini}. \citet{yuan2024mora} even designs a multi-agent framework that incorporates several advanced visual agents to replicate generalist video generation demonstrated by Sora~\citep{videoworldsimulators2024}. The advantages of collaborative AI systems are prominent. Dynamic systems make it possible to fetch external knowledge with components such as retrievers. Besides, extra modules such as verifier make the results controllable and trustworthy.

Compared to monolithic models, collaborative AI systems have posed new challenges in design and optimization. Novel paradigms are emerging to tackle these challenges, of which the most popular is composition frameworks, which build applications out of calls to models and other components~\citep{langchain2023applications,hong2023metagpt,wu2023autogen,chen2023agentverse,liang2023encouraging}. DSPy~\citep{khattab2023dspy} automatically optimizes the prompt instructions, few-shot examples and other parameters in the pipeline to maximize the end-to-end performance. GPTSwarm~\citep{zhuge2024gptswarm} supports building agents in the form of graphs and optimizing the parameters and connections as an entire system.
\section{ComfyBench}
\label{section:comfybench}

We propose ComfyBench, a new benchmark for evaluating the capability of LLM-based agents to autonomously design collaborative AI systems within the ComfyUI~\citep{comfyanonymous2023comfyui} platform. 
Given the task instruction, ComfyBench requires agents to generate workflows to describe collaborative AI systems in ComfyUI, producing the desired results after execution.

\subsection{ComfyUI Platform}
\label{section:comfyui_platform}

\noindent \textbf{Overview.} ComfyUI is an open-source platform for designing and executing generative pipelines, where users can design their workflows by creating and connecting a series of nodes. Each node stands for a module, which can be a model, a tool, or a widget. On the user side, the workflows are visualized as directed acyclic graphs (DAGs), so that users can operate them intuitively. On the server side, however, the workflows are saved in a JSON object, which describes the nodes and connectivity in a structured manner. ComfyUI executes the JSON format workflows at the granularity of nodes, automatically loading models, conducting inference, passing variables, and returning results to users.

\noindent \textbf{Features.} ComfyUI provides native support for common models and tools related to Stable Diffusion~\citep{rombach2022high}~\citep{rombach2022high}, including but not limited to VAEs~\citep{kingma2013auto}, CLIPs~\citep{radford2021learning}, LoRAs~\citep{ryu2023low}, IP-Adapters~\citep{ye2023ip}, and ControlNets~\citep{zhang2023adding}. With the joint efforts of the open-source community, hundreds of ComfyUI extensions are developed to support various latest models and tools, such as InsightFace~\citep{guo2023insightface}, IC-Light~\citep{zhang2024iclight}, and Segment Anything~\citep{kirillov2023segany}, which bring about immense potential for realizing flexible and controllable generation tasks. By designing intricate workflows, users can produce fantastic artwork that cannot be achieved by common tools.

\noindent \textbf{Limitations.} Despite the fancy effects that ComfyUI can achieve, the workflows are always manually crafted, which can be quite expensive. On the one hand, expert-level skills are required to figure out the appropriate nodes and parameters that can produce the desired results. On the other hand, workflows are designed to solve a specific task, while creating new workflows can be laborious and time-consuming.

\subsection{Benchmark Contents}
\label{section:benchmark_contents}

\begin{figure}[ht]
    \centering
    \includegraphics[width=1\linewidth]{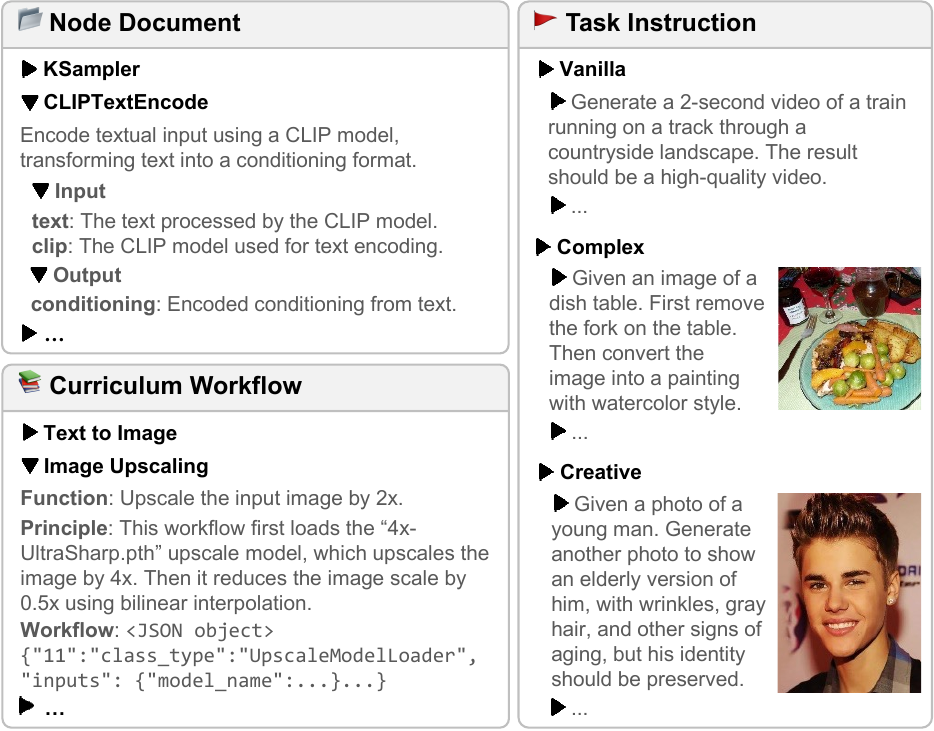}
    \vspace{-1em}
    \caption{ComfyBench provides annotations for 3205 nodes and 20 workflows, together with 200 task instructions categorized into three difficulty levels: \textit{vanilla}, \textit{complex}, and \textit{creative}.}
    \vspace{-1em}
    \label{figure:benchmark_contents}
\end{figure}

ComfyBench collects annotations for $3205$ nodes and $20$ workflows and construct $200$ tasks as is shown in Figure~\ref{figure:benchmark_contents}. More examples can be found in the supplementary material.

\noindent \textbf{Node document} provides complete documentation for the $3205$ nodes registered in ComfyUI until September 2024, serving as an important manual for agents to utilize the nodes properly. The source code is cloned from Salt AI Documentation~\citep{saltai2024salt}, which is then cleaned and arranged into documentation, describing the usage of each node and their inputs and outputs. This structured documentation serves as a solid reference resource for understanding the complex node ecosystem within the ComfyUI framework.

\noindent \textbf{Curriculum workflow} is a set of $20$ tutorial workflows in ComfyUI, which introduces some common nodes and routines to build workflows for solving basic tasks. All the workflows are verified feasible with common nodes and models in ComfyUI and correctly annotated with their purposes and principles. The curriculum workflow allows agents to master useful skills in a human-like manner.

\noindent \textbf{Task instruction} serves as the core part of the benchmark, containing $200$ task instructions. Each instruction defines a specific task, which should be solved by designing and executing a ComfyUI workflow. To reflect the generalization ability of agents, the tasks are manually categorized into three difficulty levels: \textit{vanilla}, \textit{complex}, and \textit{creative}.
\begin{itemize}
    \item \textbf{Vanilla} includes $100$ tasks, which can be solved by learning from a single curriculum workflow, involving minor modifications on several nodes and their parameters.
    \item \textbf{Complex} includes $60$ tasks, which necessitates learning from more than one curriculum workflow, involving functional combinations of multiple pipelines and minor adjustments on nodes and parameters.
    \item \textbf{Creative} includes $40$ tasks, which cannot be solved by directly imitating one or more curriculum workflows, requiring agents to understand the underlying principles and apply the learned skills in a novel way.
\end{itemize}
Besides, we provide an auxiliary dataset as the input for some of the tasks. If an image or video is involved in the task, it will be specified in the instruction.

\subsection{Evaluation Metrics}
\label{section:evaluation_metrics}

Different from the common benchmarks for image or video generation, we are evaluating the generated workflows, so traditional metrics are not applicable. We divide the evaluation into two progressive stages. The first stage is to check whether the generated workflows are syntactically and semantically correct, and the second stage is to check whether the generated workflows can produce the expected results in the task instructions. We design two metrics: \textit{pass rate} and \textit{resolve rate}, corresponding to the two stages respectively.

\noindent \textbf{Pass rate} measures the ratio of tasks where generated workflows are syntactically and semantically correct. In practice, each workflow is sent to the server of ComfyUI, which automatically conducts validation and execution. A task will be marked as passed only if the server finishes the execution process and returns a success message.

\noindent \textbf{Resolve rate} measures the ratio of tasks where generated workflows can produce the expected results in the task instruction. In other words, we need to automatically determine the consistency between the produced images or videos and the task requirements. In practice, we apply Visual Language Models (VLMs), where different prompt templates are written according to the input and output modalities. Specifically, we adopt GPT-4o for evaluation, as it demonstrates strong capabilities for understanding language and analyzing visual content. For images, we directly encode them into the prompt. For videos, we uniformly sample up to $10$ frames before encoding, which compresses the context length within a reasonable range. A task will be marked as resolved only if the GPT-4o model confirms that all the requirements are satisfied.

\begin{figure*}[!t]
    \centering
    \includegraphics[width=1\linewidth]{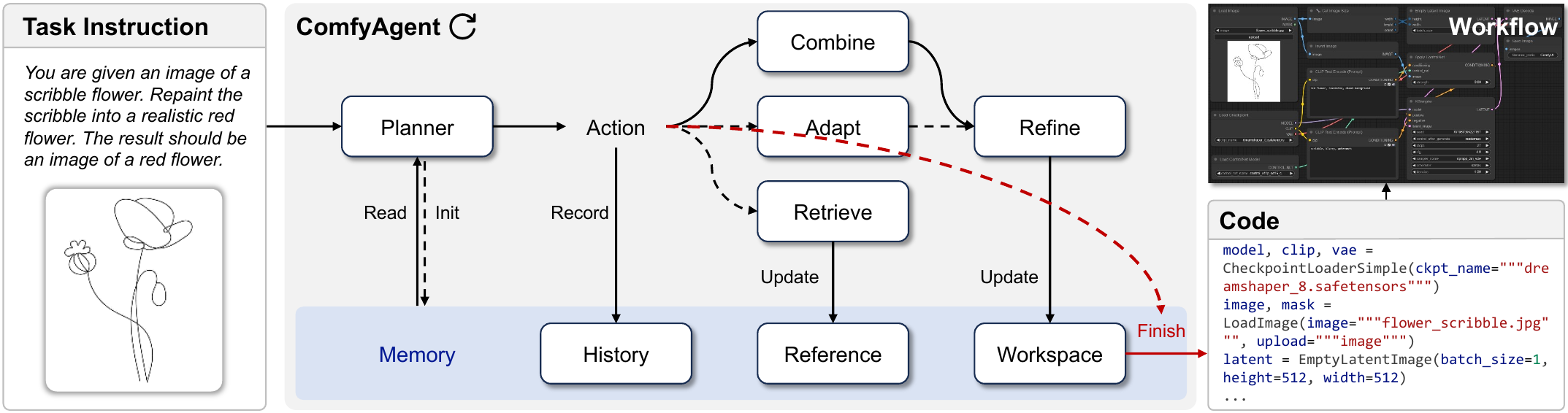}
    \caption{The architecture of the ComfyAgent framework. Multiple agents cooperate to design workflows in a step-by-step manner. Given the task instruction, the planner initializes the memory and produces a plan. For each step, the planner updates the plan and forms an action. Different actions, including \textit{combine}, \textit{adapt}, and \textit{retrieve}, are then handled by corresponding agents. After combination or adaptation, \textit{refine} action will be conducted to ensure the correctness. All the agents can interact with the memory, which consists of history, reference, and workspace. Once the task is deemed completed, the planner will finish the procedure and save the workflow.}
    \label{fig:comfyagent_framework}
\end{figure*}

\subsection{Human Evaluation}
\label{section:human_evaluation}

The computation of the resolve rate involves using VLMs to determine the consistency between the produced images or videos and the task requirements, which unavoidably involves subjectivity. To verify the reliability of VLM-based evaluation, we conduct a human evaluation to analyze the agreement between human evaluators and GPT-4o.

We randomly select $70$ tasks that are either completed by ComfyAgent proposed in Section~\ref{section:comfyagent} or by a human expert. Human evaluators and the latest GPT-4o model are asked to provide their judgments respectively. We compute the average scores for each question under a sample size of $560$ and conduct a correlation analysis. Results are presented in Table~\ref{table:human_evaluation}, where the statistics indicate a strong agreement between human evaluators and GPT-4o, which verifies the reliability of our proposed VLM-based evaluation. More details can be found in the supplementary material.

\begin{table}[ht]
    \caption{Correlation analysis between the average scores given by human evaluators and GPT-4o indicates a strong agreement.}
    \label{table:human_evaluation}
    \centering
    \begin{adjustbox}{max width=1\textwidth}
    \begin{tabular}{lcc}
        \toprule
        \textbf{Metric} & \textbf{Statistic} & \textbf{P-value} \\
        \midrule
        Kendall's $\tau$ & $0.645$ & $5.427\times 10^{-11}$ \\
        Pearson's $r$ & $0.764$ & $1.482\times 10^{-14}$ \\
        Spearman's $\rho$ & $0.737$ & $3.672\times 10^{-13}$ \\
        \bottomrule
    \end{tabular}
    \end{adjustbox}
    \vskip -0.1in
\end{table}

\section{ComfyAgent}
\label{section:comfyagent}

\subsection{Workflow Representation}
\label{section:workflow_representation}

\begin{figure}[ht]
    \centering
    \includegraphics[width=1\linewidth]{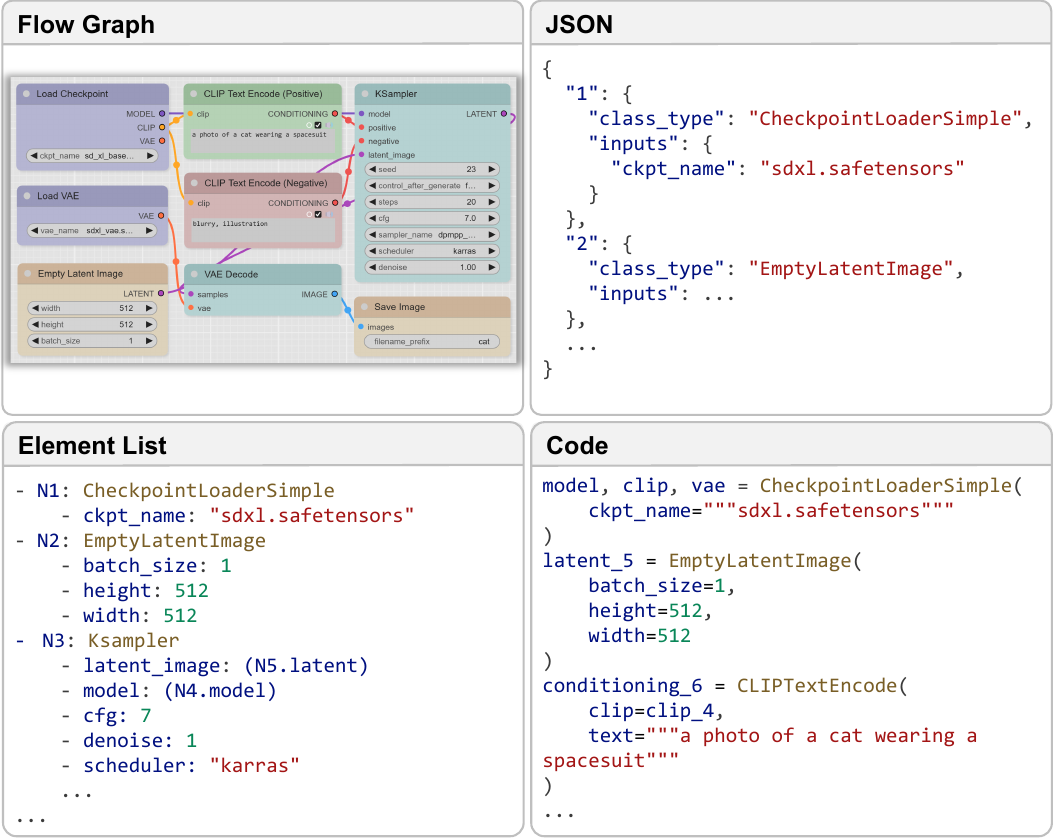}
    \caption{Examples of four common formats to represent workflows: flow graph, JSON, element list, and code.}
    \label{figure:representation}
\end{figure}

Collaborative AI systems coordinate multiple interacting components, such as AI models and external tools, using workflows. These workflows are essentially high-level diagrams that outline the operational logic. They can be naturally represented as directed acyclic graphs (DAGs) consisting of two main elements: vertices and directed edges. Vertices represent information processing components with input and output interfaces, while directed edges represent information flow between components. Specifically, the incoming edges of a vertex represent the external information it receives, while the outgoing edges signify the information it outputs, with multiple edges illustrating various information paths. This abstraction allows efficient construction and modification of collaborative AI systems, as demonstrated by platforms like ComfyUI~\citep{comfyanonymous2023comfyui}.

There are four common formats to represent workflows: flow graph, JSON, element list, and code, as shown in Figure~\ref{figure:representation}. Flow graph~\citep{comfyanonymous2023comfyui} provides intuitive visualization for humans but is unsuitable for LLM or VLM processing. JSON, typically used in~\citep{zhang2023data}, provides a structured representation but is unwieldy for complex workflows due to redundant information and LLMs' context limitations. Element list, typically used in~\citep{li2024autoflow}, is closer to natural language and provides a more compact representation, but it lacks explicit topological relationships, hindering LLMs from correctly processing complex workflows.

In contrast, code emerges as the most effective representation, offering various advantages including Turing completeness, rich semantic information, and natural compatibility with LLMs' code generation capabilities. We implement code representation using a restricted subset of Python-like syntax, involving basic operations and control structures while excluding advanced features. We also conduct an ablation study in Section~\ref{section:ablation_studies} to demonstrate the superiority of code representation over other representations.

\begin{table*}[!ht]
    \caption{Evaluation results of all the agents on ComfyBench. The average task-wise token consumption, the \textit{pass rate} and \textit{resolve rate} of every task category denoted by \%P and \%R, and the overall result on ComfyBench, are reported. The best results are highlighted in bold.}
    \label{table:evaluation_results}
    \centering
    \begin{tabular}{l|c|cccccc|cc}
        \toprule
        \multirow{2}{*}{\textbf{Agent}} & \textbf{Cost} & \multicolumn{2}{c}{\textbf{Vanilla}} & \multicolumn{2}{c}{\textbf{Complex}} & \multicolumn{2}{c|}{\textbf{Creative}} & \multicolumn{2}{c}{\textbf{Total}}\\
        \cmidrule(lr){2-2} \cmidrule(lr){3-4} \cmidrule(lr){5-6} \cmidrule(lr){7-8} \cmidrule(lr){9-10}
        & \makebox[0.1\textwidth][c]{\textbf{\#Tokens (k)}} & \makebox[0.05\textwidth][c]{\textbf{\%P}} & \makebox[0.05\textwidth][c]{\textbf{\%R}} & \makebox[0.05\textwidth][c]{\textbf{\%P}} & \makebox[0.05\textwidth][c]{\textbf{\%R}} & \makebox[0.05\textwidth][c]{\textbf{\%P}} & \makebox[0.05\textwidth][c]{\textbf{\%R}} & \makebox[0.05\textwidth][c]{\textbf{\%P}} & \makebox[0.05\textwidth][c]{\textbf{\%R}} \\
        \midrule
        GPT-4o + Zero-shot & 0.50 & 0.0 & 0.0 & 0.0 & 0.0 & 0.0 & 0.0 & 0.0 & 0.0 \\
        GPT-4o + Few-shot & 2.55 & 32.0 & 27.0 & 16.7 & 8.3 & 7.5 & 0.0 & 22.5 & 16.0 \\
        GPT-4o + CoT & 2.50 & 44.0 & 29.0 & 11.7 & 8.3 & 12.5 & 0.0 & 28.0 & 17.0 \\
        GPT-4o + CoT-SC & 11.03 & 45.0 & 34.0 & 11.7 & 5.0 & 15.0 & 0.0 & 29.0 & 18.5 \\
        Claude-3.5-Sonnet + RAG & 5.73 & 27.0 & 13.0 & 23.0 & 6.7 & 7.5 & 0.0 & 22.0 & 8.5 \\
        Llama-3.1-70B + RAG & 5.04 & 58.0 & 32.0 & 23.0 & 10.0 & 15.0 & 5.0 & 39.0 & 20.0 \\
        GPT-4o + RAG & \textbf{4.97} & 62.0 & 41.0 & 45.0 & 21.7 & \textbf{40.0} & 7.5 & 52.0 & 23.0 \\
        o1-mini + RAG & 6.05 & 32.0 & 16.0 & 21.7 & 8.3 & 12.5 & 7.5 & 25.0 & 12.0 \\
        o1-preview + RAG & 9.46 & \textbf{70.0} & \textbf{46.0} & \textbf{48.3} & \textbf{23.3} & 30.0 & 12.5 & 55.5 & \textbf{32.5} \\
        \midrule
        Llama-3.1-70B + ComfyAgent & 24.22 & 63.0 & 35.0 & 26.7 & 18.3 & 20.0 & 5.0 & 43.5 & 24.0 \\
        GPT-4o + ComfyAgent & \textbf{14.96} & 67.0 & \textbf{46.0} & \textbf{48.3} & 21.7 & \textbf{40.0} & \textbf{15.0} & \textbf{56.0} & \textbf{32.5} \\
        \bottomrule
    \end{tabular}
\end{table*}

\subsection{Multi-Agent Framework}
\label{section:multiagent_framework}

Through code representation, we can effectively prompt LLMs to generate workflows for the expected collaborative AI systems. However, with the complexity of workflows increasing, traditional agents suffer from context limitation and hallucination problems, failing to generate workflows correctly and stably, which is proved in Section~\ref{section:evaluation_results}.

To achieve reliable generation of ComfyUI workflows, we propose ComfyAgent, a multi-agent framework to design workflows in a step-by-step manner. As demonstrated in Figure~\ref{fig:comfyagent_framework}, ComfyAgent consists of three independent modules: Memory, Planner, and Actions.

\noindent \textbf{Memory} stores the recent state of ComfyAgent, which is built on psychological theories~\citep{atkinson1968human}. Specifically, it refers to the working memory~\citep{baddeley1992working}, reflecting the current circumstances, namely the recent history behaviors, results from intermediate steps, internal reasoning, and external knowledge. We formulate them into three parts.
\begin{itemize}
    \item \textbf{History} maintains recent plans and actions of Planner, enabling action review for subsequent planning.
    \item \textbf{Reference} stores information retrieved from the knowledge base, and can be updated through active retrieval.
    \item \textbf{Workspace} contains the current workflow together with its natural language annotation, which maintains a complete record of the current working status.
\end{itemize}

\noindent \textbf{Planner}, which can be specified as PlanAgent, serves as the core of ComfyAgent, providing the global scheme to design and modify workflows under the task instruction.
\begin{itemize}
    \item At the beginning of the task, PlanAgent selects an existing workflow to initialize the memory and produces a thorough multi-step plan based on the task instruction.
    \item For each step, PlanAgent produces a high-level plan, together with an action based on the current memory, which will then be processed by other agents.
    \item For each step, PlanAgent evaluates the completion status of the task. Once the task is deemed completed, PlanAgent will finish the procedure and save the workflow.
\end{itemize}

\noindent \textbf{Actions} represent different actions that can be selected by PlanAgent. All the actions are meant to modify the current memory. In ComfyAgent, we have three actions.
\begin{itemize}
    \item \textbf{Combine} is processed by CombineAgent, which aims to combine the current workflow with another workflow from references specified by PlanAgent.
    \item \textbf{Adapt} is processed by AdaptAgent, which is responsible for adapting the details (\eg parameters) of the current workflow based on the prompt given by PlanAgent.
    \item \textbf{Retrieve} is processed by RetrieveAgent, which follows the prompt given by PlanAgent to retrieve relevant information and update references accordingly.
\end{itemize}
After combination or adaptation, the updated workflow will be checked and refined by RefineAgent before entering the workspace, which avoids the accumulation of errors.

After the action is processed, ComfyAgent will enter a new step, where PlanAgent updates the existing plan and forms a new action. In such a way, ComfyAgent gradually generates the workflow. When the procedure is finished, the code representation will be converted into the standard format, which exactly describes the collaborative AI system.

\section{Experiments}
\label{section:experiments}

\subsection{Baseline Agents}
\label{section:baseline_agents}

ComfyBench places strong requirements on the learning and generalization ability of agents, necessitating LLMs with powerful reasoning capabilities and sufficient context length. Therefore, we conduct experiments on the advanced models, involving both commercial and open-source LLMs.
\begin{itemize}
    \item \textbf{Llama-3.1} (\texttt{llama-3.1-70b-instruct})~\citep{dubey2024llama3} is an LLM developed by Meta, achieving state-of-the-art performance in the open-source community.
    \item \textbf{Claude-3.5} (\texttt{claude-3.5-sonnet-20240620})~\citep{anthropic2024claude} is a commercial LLM developed by Anthropic, with a reputation for its creativity and interaction quality.
    \item \textbf{GPT-4o} (\texttt{gpt-4o-2024-08-06})~\citep{openai2024hello} is currently the mainstream model provided by OpenAI and most widely used by the community. It still maintains state-of-the-art performance in leaderboards of various domains.
    \item \textbf{o1-mini} (\texttt{o1-mini-2024-09-12}) and \textbf{o1-preview} (\texttt{o1-preview-2024-09-12})~\citep{openai2024introducing} are the latest models released by OpenAI, which focus on complex reasoning tasks such as mathematics and programming.
\end{itemize}
We adopt five methods that have proved to be universally effective by the community and can be conveniently adapted to solve the tasks in ComfyBench.
\begin{itemize}
    \item \textbf{Zero-shot Learning} directly feeds LLMs with the task instruction to conduct inference.
    \item \textbf{Few-shot Learning}~\citep{brown2020language} provides a set of demonstrations in the prompt, which utilizes the in-context learning ability of LLMs and facilitates the correct generation.
    \item \textbf{Chain-of-Thought (CoT)}~\citep{wei2022chain} is improved based on few-shot learning, which instructs the agent to articulate the reasoning process before providing the final answer.
    \item \textbf{CoT with Self-consistency (CoT-SC)}~\citep{wang2022self} is an improved version of CoT, which ensembles parallel trajectories and then selects the most consistent answer.
    \item \textbf{Retrieval-Augmented Generation (RAG)}~\citep{lewis2020retrieval} learns from the retrieved demonstrations, which efficiently utilizes the most relevant information.
\end{itemize}
Considering the limited experimental budget, we evaluate the methods mainly on GPT-4o. ComfyAgent is also evaluated on Llama-3.1 to verify its performance on the open-source model. RAG is evaluated on all the models to provide a horizontal comparison of their capabilities.

\subsection{Implementation Details}
\label{section:implementation_details}

\noindent \textbf{ComfyUI Settings.} The experiments are conducted on a fixed version of ComfyUI with necessary extensions. A set of common models is installed to make the workflows runnable and the results reproducible. In ComfyUI, the workflows are natively represented in JSON format. We implement a parser to extract the entire DAG described in the JSON object, which is then converted into a sequence of Python-like function calls in topological order. Another parser is implemented to restore this conversion. This ensures that workflows can be effectively generated by agents through code representation while remaining fully compatible with the ComfyUI execution environment. Note that the node documents are not fed into the agents due to context limitations, even though this technique would potentially improve their performance.

\noindent \textbf{Agent Parameters.} We follow CodeAct~\citep{wang2024executable} to format the interaction with LLMs. The parameters of the agents are manually set to exploit their largest potential within a limited context length. For few-shot learning, CoT, and CoT-SC, the number of demonstrations is consistently set to $3$, but CoT-SC will ensemble $3$ trajectories in parallel to form the final answer. For RAG and ComfyAgent, we adopt an advanced embedding model (\texttt{text-embedding-3-large})~\citep{openai2024new} released by OpenAI to encode the texts for retrieval, and the number of retrieved demonstrations is set to $5$. For ComfyAgent, the modules are implemented as different roles played by the same LLM. We provide the prompt implementation and an example trajectory in the supplementary material.

\subsection{Evaluation Results}
\label{section:evaluation_results}

\begin{table}[!t]
    \caption{Ablation results of RAG on GPT-4o with different workflow representations on ComfyBench. The \textit{pass rate} and \textit{resolve rate} are reported, denoted by \%P and \%R respectively.}
    \label{table:ablation_representation}
    \small
    \centering
    \begin{adjustbox}{max width=1.\textwidth}
    \begin{tabular}{lcccccc}
        \toprule
        \multirow{2}{*}{\textbf{Representation}} & \multicolumn{2}{c}{\textbf{Vanilla}} & \multicolumn{2}{c}{\textbf{Complex}} & \multicolumn{2}{c}{\textbf{Creative}} \\
        \cmidrule(lr){2-3} \cmidrule(lr){4-5} \cmidrule(lr){6-7}
        & \textbf{\%P} & \textbf{\%R} & \textbf{\%P} & \textbf{\%R} & \textbf{\%P} & \textbf{\%R} \\
        \midrule
        Flow Graph & - & - & - & - & - & - \\
        JSON & 52.0 & 33.0 & 25.0 & 8.3 & 17.5 & 7.5 \\
        Element List & 51.0 & 30.0 & 23.3 & 5.0 & 20.0 & \textbf{10.0} \\
        Code & \textbf{62.0} & \textbf{41.0} & \textbf{45.0} & \textbf{15.0} & \textbf{30.0} & 7.5 \\
        \bottomrule
    \end{tabular}
    \end{adjustbox}
\end{table}

\begin{table}[!t]
    \caption{Ablation results of ComfyAgent on GPT-4o with different architectures on ComfyBench. Each variant is implemented by removing one agent from the multi-agent framework. The \textit{pass rate} and \textit{resolve rate} are reported, denoted by \%P and \%R respectively.}
    \label{table:ablation_architecture}
    \small
    \centering
    \begin{adjustbox}{max width=1.\textwidth}
    \begin{tabular}{lcccccc}
        \toprule
        \multirow{2}{*}{\textbf{Architecture}} & \multicolumn{2}{c}{\textbf{Vanilla}} & \multicolumn{2}{c}{\textbf{Complex}} & \multicolumn{2}{c}{\textbf{Creative}} \\
        \cmidrule(lr){2-3} \cmidrule(lr){4-5} \cmidrule(lr){6-7}
        & \textbf{\%P} & \textbf{\%R} & \textbf{\%P} & \textbf{\%R} & \textbf{\%P} & \textbf{\%R} \\
        \midrule
        ComfyAgent & 67.0 & \textbf{46.0} & \textbf{48.3} & \textbf{21.7} & \textbf{40.0} & \textbf{15.0} \\
        w/o Combine & \textbf{69.0} & 43.0 & 15.0 & 5.0 & 35.0 & \textbf{15.0} \\
        w/o Adapt & 52.0 & 26.0 & 20.0 & 10.0 & 17.5 & 5.0 \\
        w/o Retrieve & 27.0 & 16.0 & 20.0 & 5.0 & 7.5 & 7.5 \\
        w/o Refine & 65.0 & 42.0 & 43.3 & \textbf{21.7} & 35.0 & 12.5 \\
        \bottomrule
    \end{tabular}
    \end{adjustbox}
\end{table}     

\begin{table*}[!t]
    \caption{Some examples generated by ComfyAgent on ComfyBench. Given the task instruction, ComfyAgent designs a workflow to describe the collaborative AI system in ComfyUI, which is then executed to generate the result required by the instruction.}
    \label{table:generation_examples}
    \centering
    \begin{tabular}{c@{\hskip 4pt}c|c@{\hskip 4pt}c|c@{\hskip 4pt}c}
        \toprule
        \multicolumn{2}{c|}{\small{\textbf{Task Instruction}}} & \multicolumn{2}{c|}{\small{\textbf{Task Instruction}}} & \multicolumn{2}{c}{\small{\textbf{Task Instruction}}} \\ [2pt]
        \multicolumn{2}{p{0.3\linewidth}|}{\small{\textit{You are given an image of a male celebrity. Transform the man in the image into a beautiful woman with ponytail hair while preserving her facial identity.}}} & \multicolumn{2}{p{0.3\linewidth}|}{\small{\textit{You are given an image of a standing cat. Replace the background with a scene of a cozy living room while keeping the lighting and shadows consistent.}}} & \multicolumn{2}{p{0.3\linewidth}}{\small{\textit{You are given an image of a red bridge with a person standing on it. Remove the person from the image while maintaining the original appearance of the bridge.}}} \\ [38pt]
        \multicolumn{1}{c}{\small{\textbf{Input}}} & \multicolumn{1}{c|}{\hspace{-6pt}\small{\textbf{Result}}} & \multicolumn{1}{c}{\small{\textbf{Input}}} & \multicolumn{1}{c|}{\hspace{-6pt}\small{\textbf{Result}}} & \multicolumn{1}{c}{\small{\textbf{Input}}} & \multicolumn{1}{c}{\hspace{-6pt}\small{\textbf{Result}}} \\ [2pt]
        \includegraphics[width=0.15\linewidth]{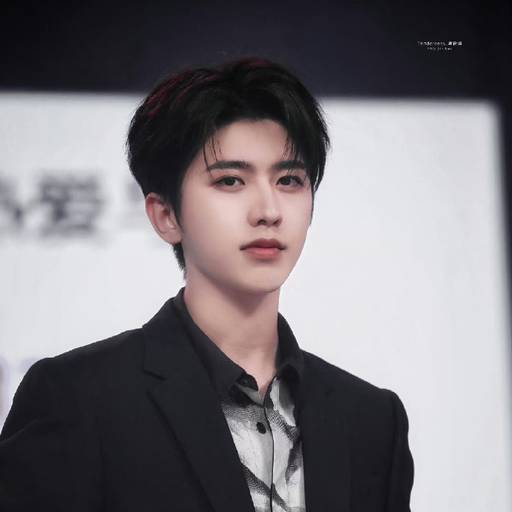} & \includegraphics[width=0.15\linewidth]{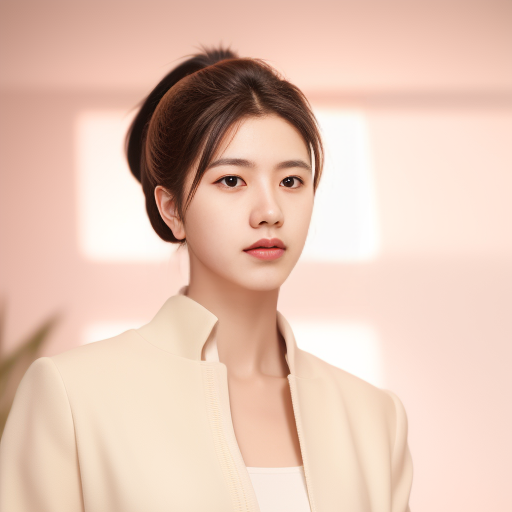} & \includegraphics[width=0.15\linewidth]{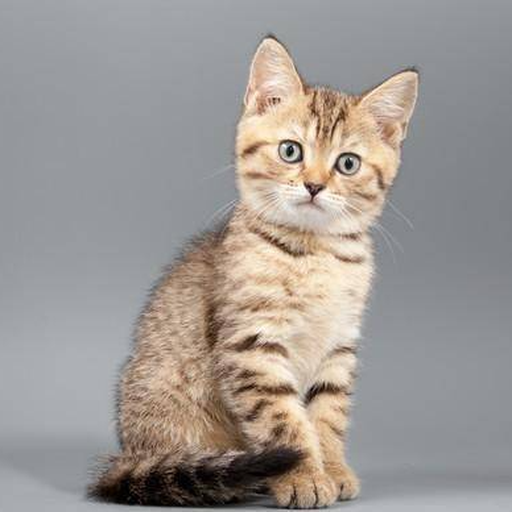} & \includegraphics[width=0.15\linewidth]{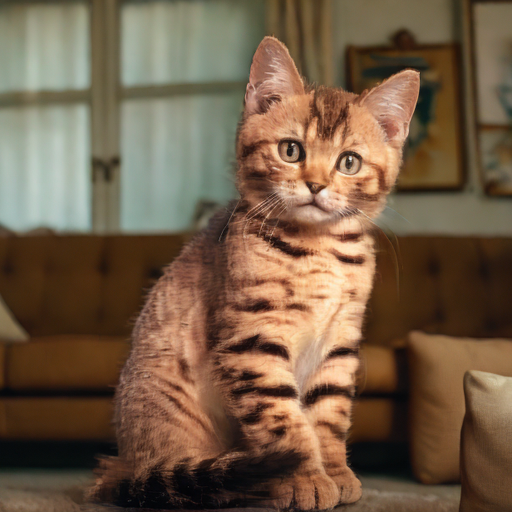} & \includegraphics[width=0.15\linewidth]{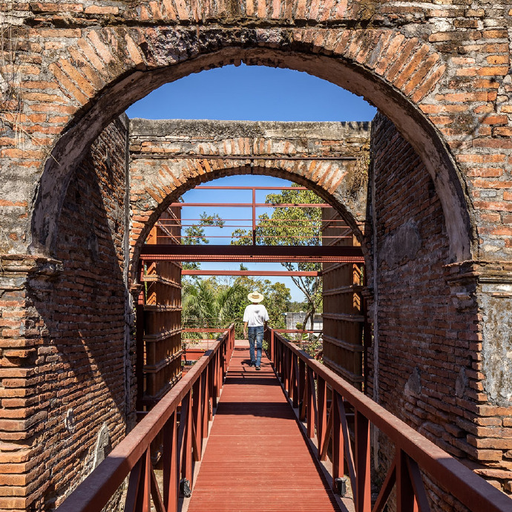} & \includegraphics[width=0.15\linewidth]{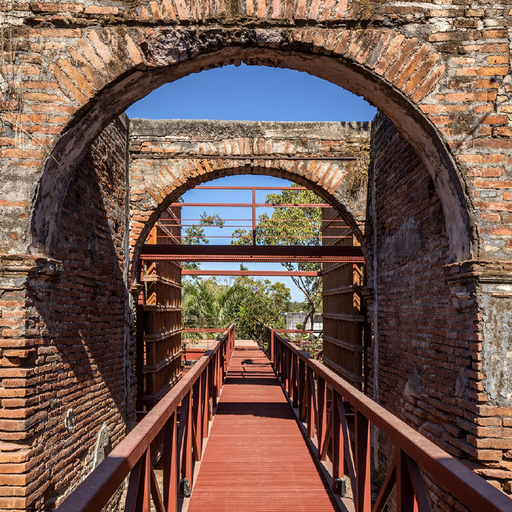} \\
        \bottomrule
    \end{tabular}
\end{table*}

We report the evaluation results of different agents on ComfyBench in Table~\ref{table:evaluation_results}. All the agents follow the parameter settings in Section~\ref{section:implementation_details} and adopt code representation in Section~\ref{section:workflow_representation}. We also present some examples generated by ComfyAgent on ComfyBench in Table~\ref{table:generation_examples}. More examples can be found in the supplementary material.

In terms of methods, zero-shot learning fails due to a lack of domain knowledge. CoT achieves a better performance than few-shot learning, and CoT-SC shows a slight improvement over CoT. However, these agents fail to achieve comparable performance with retrieval-based agents, because they cannot fully exploit the external knowledge provided by ComfyBench. ComfyAgent outperforms all the baselines in every task category, achieving an overall pass rate of $56\%$ and resolve rate of $32.5\%$, which indicates that the multi-agent framework applied in ComfyAgent is capable of alleviating the hallucination problem and improving the generation quality and stability. This improvement is consistent on both open-source and commercial models.

In terms of models, Claude-3.5 and o1-mini show poor performance under RAG, which is mainly caused by the hallucination problem, where models modify the nodes or parameters without verifying their validity. In contrast, Llama-3.1 and GPT-4o show less hallucination and achieve better performance, and o1-preview achieves an outstanding performance comparable to ComfyAgent on GPT-4o.

\subsection{Ablation Studies}
\label{section:ablation_studies}

\noindent \textbf{Representation.} We conduct an ablation study to verify the advantages of representing workflows with code introduced in Section~\ref{section:workflow_representation}. We implement three variants of RAG on GPT-4o, where the workflows are respectively represented in JSON, element list, and code. The variants are evaluated on ComfyBench with the same parameter settings as in Section~\ref{section:evaluation_results}. Flow graph is not implemented because it is not understandable by LLM-based agents. We report the ablation results in Table~\ref{table:ablation_representation}, where RAG with code representation shows superior performance over other representations, which verifies our claim. RAG with JSON representation is expected to perform even worse without the unique feature of GPT-4o to generate structured JSON objects.

\noindent \textbf{Architecture.} We conduct another ablation to demonstrate the rationality of the multi-agent framework introduced in Section~\ref{section:multiagent_framework}, where four variants of ComfyAgent on GPT-4o are implemented, each removing one agent from the original framework. We report the ablation results in Table~\ref{table:ablation_architecture}. We find that removing any of the four agents will lead to a performance drop. Removing CombineAgent alleviates hallucination in the vanilla category, but leads to failure in the complex category, demonstrating the necessity of integrating multiple demonstrations. Removing AdaptAgent or RetrieveAgent leads to a significant performance drop in all the categories, which underscores the critical role of detail modification and knowledge utilization. Removing RefineAgent leads to a minor but noticeable performance drop in all the categories, indicating its effectiveness in mitigating error accumulation. Therefore, we can conclude that our meticulously-designed multi-agent framework contributes to the superior performance of ComfyAgent.
\section{Conclusion}
\label{section:conclusion}

In this work, we investigate the potential of LLM-based agents to autonomously design collaborative AI systems. Building upon the open-source ComfyUI~\citep{comfyanonymous2023comfyui} platform, we present ComfyBench, an autonomous workflow design environment serving as both a benchmark for evaluating the capabilities of agents in designing collaborative systems and a means to extend the applicability of ComfyUI in real-world, language-instruction-based generative applications. We further develop ComfyAgent, which achieves a comparable resolve rate to o1-preview~\citep{openai2024introducing} and significantly surpasses other agents on ComfyBench. We point out two key insights underpinning ComfyAgent. First, representing workflows through code is efficient, which provides rich semantics for agents to better understand complex systems. Second, the multi-agent framework of ComfyAgent highlights the substantial benefits of task decomposition and collaborative generation. However, the top-performing agent, ComfyAgent, has achieved a resolve rate of only 15\% for creative tasks on ComfyBench. LLM-based agents still have a long way to go in designing collaborative AI systems. We hope that ComfyBench and ComfyAgent can serve as valuable resources in advancing the development of intelligent and autonomous collaborative AI systems.

\noindent\textbf{Limitations.} Well-designed fine-tuning methods could potentially elevate agents to an expert level on ComfyBench, but they are quite expensive and thus not involved in our work. In addition, ComfyAgent operates in an open-loop manner. An alternative approach would involve using VLMs as evaluators to provide feedback~\citep{hu2024automated}, enabling a closed-loop system where agents can refine their workflows based on evaluations, thus leading to better results.

\noindent\textbf{Acknowledgments.} This work is supported by the Shanghai Municipal Science and Technology Major Project.
{
    \small
    \bibliographystyle{ieeenat_fullname}
    \bibliography{main}
}

\clearpage
\setcounter{page}{1}
\maketitlesupplementary
\appendix

\section*{Overview}
This supplementary document provides additional details to support our main manuscript, organized as follows:
\begin{itemize}
    \item Section~\ref{appendix:more_task_instructions} presents more examples of task instructions that are included in ComfyBench.
    \item Section~\ref{appendix:more_comfyagent_examples} provides more examples generated by ComfyAgent on ComfyBench, as well as a typical trajectory of ComfyAgent to solve a task on ComfyBench.
    \item Section~\ref{appendix:vlm_based_evaluation_implementation} summarizes the detailed implementation of GPT-4o evaluaiton for resolve rate computation.
    \item Section~\ref{appendix:details_for_human_evaluation} demonstrates the details of human evaluation, certifying the stability of our VLM evaluation system.
    \item Section~\ref{appendix:comfyagent_prompt_implementation} showcases the detailed prompt implementation for each agent in the ComfyAgent framework.
    \item Section~\ref{appendix:details_for_inference_parameters} provides the details of inference parameters.
    \item Section~\ref{appendix:metrics_for_generation_quality} presents some metrics on the generation quality of ComfyAgent compared with single-model methods.
    \item Section~\ref{appendix:comparison_with_multi_agent_methods} presents the performance comparison between ComfyAgent and some common multi-agent methods.
\end{itemize}

\section{More Task Instructions}
\label{appendix:more_task_instructions}

In Table~\ref{table:task_instruction}, we present more examples of task instruction to provide a deeper preview of ComfyBench. The modalities and categories of the tasks are also included. ComfyBench covers a wide range of image and video generation tasks. Each instruction describes an expected result to realize. The tasks are categorized into three difficulty levels: \textit{vanilla}, \textit{complex}, and \textit{creative}, which reflect the generalization capability of LLM-based agents.

\section{More ComfyAgent Examples}
\label{appendix:more_comfyagent_examples}

As an extension, we present more examples generated by ComfyAgent on ComfyBench in Table~\ref{table:more_examples} to demonstrate the fantastic effects that ComfyAgent can achieve. Restricted by the format, we only present the image examples. Complete examples, including those involving videos as inputs or results, are available on our project website: \href{https://xxyqwq.github.io/ComfyBench}{https://xxyqwq.github.io/ComfyBench}.

Furthermore, we present a typical example trajectory of ComfyAgent in Figure~\ref{figure:example_trajectory}, which specifically demonstrates how ComfyAgent gradually designs a complete workflow for the collaborative AI system according to the task instruction. The task is selected from the complex category in ComfyBench, which requires generating a video and interpolating the frames, involving common techniques such as text-to-video generation and video interpolation.

\section{VLM-based Evaluation Implementation}
\label{appendix:vlm_based_evaluation_implementation}

In ComfyBench, we adopt the latest GPT-4o model to compute the resolve rate. We design prompt templates for each task modality, where the images and videos can be encoded. Considering GPT-4o cannot directly process videos, we uniformly sample up to $10$ frames and input them as a sequence of images, so that the context length can be controlled within a reasonable range. Since the tasks in ComfyBench involve operations such as upscaling and interpolation, we also provide the model with the original resolution and frame rate information. The model must conduct the necessary analysis before providing the result so that the judgment can be more reasonable and reliable.

Specifically, we present a typical example of VLM-based evaluation in Figure~\ref{figure:vlmbased_evaluation}, which evaluates a text-to-image generation task. The answer given by GPT-4o involves a thorough analysis of the consistency between the task requirements and the generated image, followed by a correct final judgment, demonstrating a strong capability to understand the points and determine the consistency.

\section{Details for Human Evaluation}
\label{appendix:details_for_human_evaluation}

\begin{figure}[ht]
    \centering
    \includegraphics[width=0.8\linewidth]{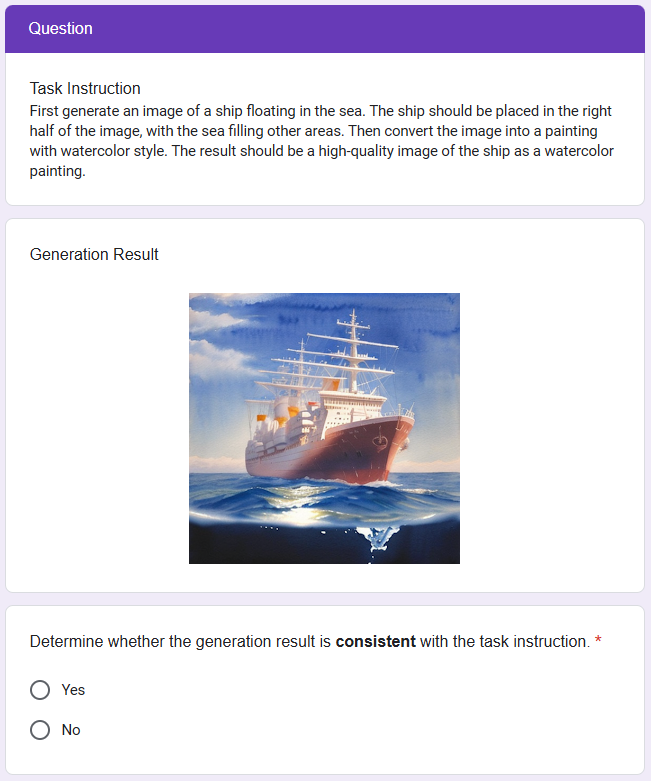}
    \caption{A sample question selected from the created questionnaires on Google Forms in the human evaluation.}
    \label{figure:sample_question}
\end{figure}

We select $50$ tasks from ComfyBench that are completed by ComfyAgent. For the rigor of the human evaluation, we choose additional $20$ tasks from ComfyBench that are not completed by ComfyAgent and manually construct their generation results. Therefore, a total of $70$ questions are formed finally. In each question, we provide the task instruction, the input image or video (if any), and the generated image or video, where the answer should be either ``Yes'' or ``No'', representing whether the generated result is consistent with the task instruction. We create multiple questionnaires in Google Forms for distribution, each containing $20$ questions. A sample question selected from the questionnaires is presented in Figure~\ref{figure:sample_question}.

We collect $28$ answer sheets from human evaluators to form a sample size of $560$, with each question answered by $8$ human evaluators. We also prompt the latest GPT-4o model to sample $8$ answers for each question. All the answers are viewed as $0/1$ variables to compute the average scores given by human evaluators and GPT-4o respectively, which indicate their tendency on each question.

\begin{figure}[ht]
    \centering
    \includegraphics[width=1\linewidth]{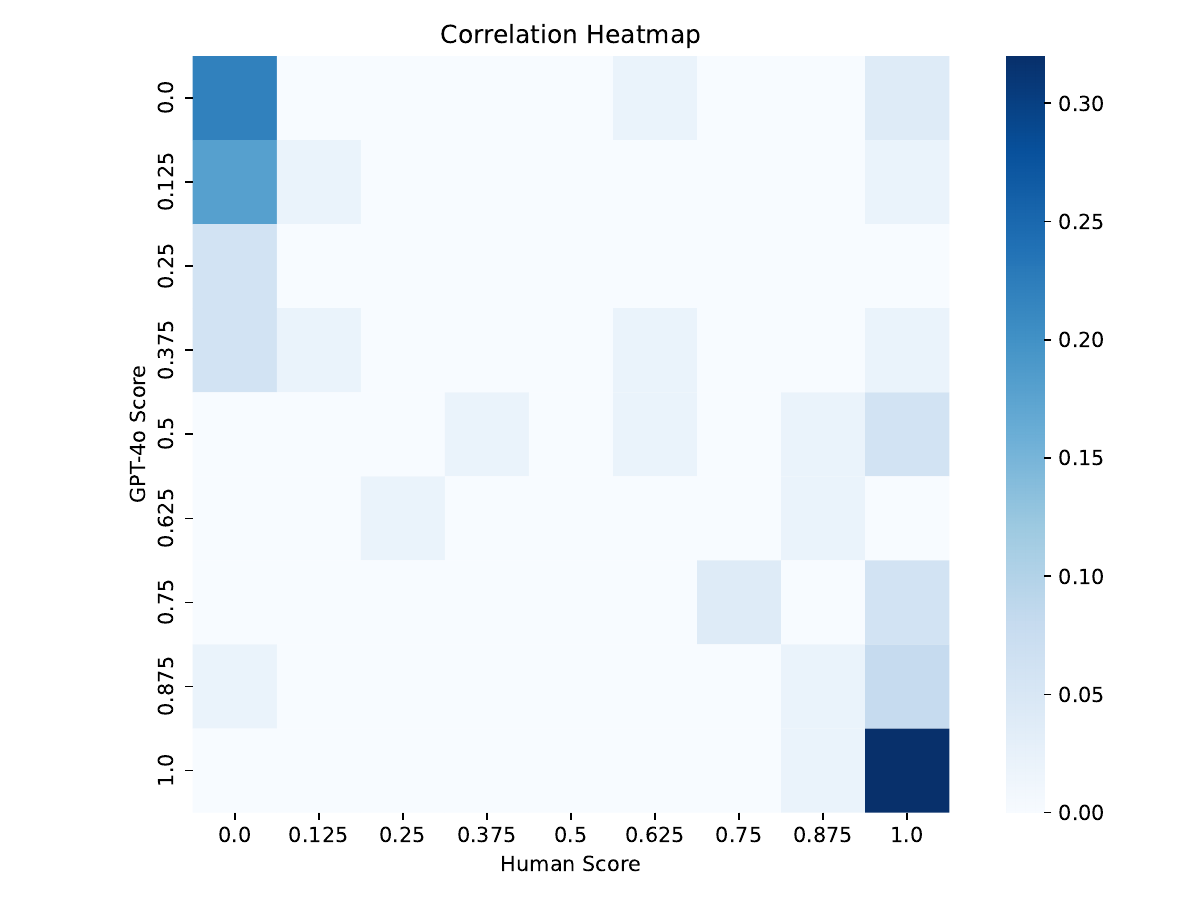}
    \caption{Correlation heatmap of the scores given by human evaluators and GPT-4o in the human evaluation.}
    \label{figure:correlation_heatmap}
\end{figure}

Based on the $70$ pairs of average scores, we present a heatmap in Figure~\ref{figure:correlation_heatmap} to intuitively demonstrate the correlation between the scores given by human evaluators and GPT-4o. We also calculate Kendall's $\tau$, Pearson's $r$, and Spearman's $\rho$. Both the heatmap and the statistics indicate a strong agreement between human evaluators and GPT-4o.

\section{ComfyAgent Prompt Implementation}
\label{appendix:comfyagent_prompt_implementation}

We provide the prompt implementation of ComfyAgent, including the prompt templates for PlanAgent, CombineAgent, AdaptAgent, and RefineAgent, which are respectively presented in Figure~\ref{figure:planagent_prompt},~\ref{figure:combineagent_prompt},~\ref{figure:adaptagent_prompt}, and~\ref{figure:refineagent_prompt}. Generally, we first introduce the background information of the ComfyUI platform, then present the necessary information for the current task, and finally specify the answer format. We follow the prompting strategy in CodeAct, which encloses answers with XML tags, to extract the results. The memory and RetrieveAgent are implemented as rule-based modules, so prompt templates are not available for them.

\section{Details for Inference Parameters}
\label{appendix:details_for_inference_parameters}

Taking into account the variations among different LLMs, we retain most parameters, such as context length and probability threshold, at their default settings, allowing each model to fully utilize its capabilities. Given that the implementation of various agents relies on multiple samplings, the temperature parameter is set to $1.0$ to balance reliability with creativity. Meanwhile, for stable VLM-based evaluations, the temperature parameter is fixed at $0$.

\section{Metrics for Generation Quality}
\label{appendix:metrics_for_generation_quality}

We use DALL-E 3 and InstructPix2Pix as baseline single-model methods for image generation and editing respectively. ComfyAgent completes 112 out of 200 tasks on ComfyBench, while DALL-E 3 and InstructPix2Pix only complete 44 and 20 tasks respectively. The HPSv2 and Aesthetic Score on the available results are shown in Table~\ref{table:generation_quality}, demonstrating that ComfyAgent can achieve similar generation quality compared to common single-model methods.

\begin{table}[ht]
    \caption{Comparison with common single-model methods.}
    \label{table:generation_quality}
    \small
    \centering
    \begin{tabular}{lcccc}
        \toprule
        \multirow{2}{*}{\textbf{Method}} & \multicolumn{2}{c}{\textbf{Image Generation}} & \multicolumn{2}{c}{\textbf{Image Editing}} \\
        \cmidrule(lr){2-3} \cmidrule(lr){4-5}
        & HPSv2 & Aesthetic & HPSv2 & Aesthetic \\
        \midrule
        DALL-E 3 & 0.299 & 6.212 & - & - \\
        InstructPix2Pix & - & - & 0.225 & 5.078 \\
        ComfyAgent & 0.289 & 6.353 & 0.197 & 4.976 \\
        \bottomrule
    \end{tabular}
\end{table}

\section{Comparison with Multi-agent Methods}
\label{appendix:comparison_with_multi_agent_methods}

We select CAMEL and Multi-agent Debate (MAD) as baseline multi-agent approaches, as they can be seamlessly adapted to address the tasks in ComfyBench. The evaluation results, summarized in Table~\ref{table:multi_agent_methods_comparison}, demonstrate that ComfyAgent significantly outperforms these baseline methods, underscoring its effectiveness in autonomously designing collaborative AI systems.

\begin{table}[ht]
    \caption{Comparison with other multi-agent frameworks.}
    \label{table:multi_agent_methods_comparison}
    \small
    \centering
    \begin{tabular}{lcccccc}
        \toprule
        \multirow{2}{*}{\textbf{Method}} & \multicolumn{2}{c}{\textbf{Vanilla}} & \multicolumn{2}{c}{\textbf{Complex}} & \multicolumn{2}{c}{\textbf{Creative}} \\
        \cmidrule(lr){2-3} \cmidrule(lr){4-5} \cmidrule(lr){6-7}
        & \%P & \%R & \%P & \%R & \%P & \%R \\
        \midrule
        CAMEL & 23.0 & 14.0 & 13.3 & 3.3 & 5.0 & 0.0 \\
        MAD & 49.0 & 23.0 & 28.3 & 5.0 & 30.0 & 2.5 \\
        ComfyAgent & \textbf{67.0} & \textbf{46.0} & \textbf{48.3} & \textbf{21.7} & \textbf{40.0} & \textbf{15.0} \\
        \bottomrule
    \end{tabular}
\end{table}


\clearpage

\begin{table*}
    \caption{More examples of task instruction in ComfyBench. We present the instructions together with their modalities and categories.}
    \label{table:task_instruction}
    \centering
    \begin{tabular}{p{0.7\linewidth}cc}
        \toprule
        \textbf{Task Instruction} & \textbf{Modality} & \textbf{Category} \\
        \midrule
        \textit{Generate an image of a hotel room containing a bed, a desk, and a window. The result should be a high-quality image.} & T2I & Vanilla \\
        \midrule
        \textit{You are given an image ``many\_people.png'', which is a photo of a crowd of people. Create a 2-second video of the people cheering based on the image. The result should be a high-quality video.} & I2V & Vanilla \\
        \midrule
        \textit{You are given an image ``street\_car.png'' of a car parked on the street. Replace the red car with a green one. The result should be a high-quality image without visible artifacts.} & I2I & Vanilla \\
        \midrule
        \textit{You are given a video ``passing\_car.mp4'' of a gray car passing by on the road. Interpolate the video to increase the frame rate by 2x. The result should be a smoother video of the gray car passing by.} & V2V & Vanilla \\
        \midrule
        \textit{First generate an image of a city street at night. Then upscale it by 2x. The result should be a high-resolution image of a city street.} & T2I & Complex \\
        \midrule
        \textit{First generate a 2-second video of a bustling city at night with skyscrapers and bright lights. Then interpolate the video to increase the frame rate by 3x. The result should be a smoother video of the city at night.} & T2V & Complex \\
        \midrule
        \textit{You are given an image ``mountain\_stream.png'' of a stream flowing through a mountain. First remove the train near the stream. Then convert the image into a painting with watercolor style. The result should be a high-quality image of the stream as a watercolor painting.} & I2I & Complex \\
        \midrule
        \textit{You are given an image ``woman\_photo.jpg'', which is a photo of a woman smiling. First convert it into a portrait while keeping other details. Then replace the background with a scene of a night street. The result should be an image of the woman as a portrait in a night street.} & I2I & Complex \\
        \midrule
        \textit{Generate a poster for a series named ``Breaking Bad''. The result should be a vertical image of the main character, Walter White, wearing a white vest with a serious expression, standing in front of a motorhome in the desert. The title ``Breaking Bad'' should be displayed in the top center of the image, followed by the tagline ``Say my name''.} & T2I & Creative \\
        \midrule
        \textit{You are given an image ``cosmetic\_product.jpg'', which contains two bottles of cosmetic products illuminated by a soft yellow light. Modify the illumination into a bright pink light to create a more vibrant and attractive appearance. The result should be an image of the cosmetic products with the new illumination.} & I2I & Creative \\
        \midrule
        \textit{You are given an image ``warm\_bedroom.jpg'', which is a photo of a bedroom including a bed, a chair and some decorations. Generate a 3-second video based on the image to show the panoramic view of the bedroom from different angles. The result should be a video that explores the bedroom in a smooth manner.} & T2V & Creative \\
        \midrule
        \textit{You are given a video ``male\_idol.mp4'' of a male idol dancing in a room. Convert the video into a sketch-style animation with black strokes and white background. The result should be a sketch-style video of the idol dancing while maintaining the original elements and movements.} & V2V & Creative \\
        \bottomrule
    \end{tabular}
\end{table*}

\clearpage

\begin{table*}
    \caption{More examples produced with the collaborative AI systems designed by ComfyAgent on ComfyBench. We present the task instructions, generated results, along with image or video inputs if they are required in the task.}
    \label{table:more_examples}
    \centering
    \begin{tabular}{p{0.6\linewidth}cc}
        \toprule
        \textbf{Task Instruction} & \textbf{Input} & \textbf{Result} \\
        \midrule
        \begin{minipage}{1\linewidth}
            \textit{Generate an image of a hot air balloon floating over a scenic valley at sunrise. The result should be a high-quality image.}
        \end{minipage} & \raisebox{-0.5\height}{N/A} & \raisebox{-0.5\height}{\includegraphics[width=0.15\linewidth]{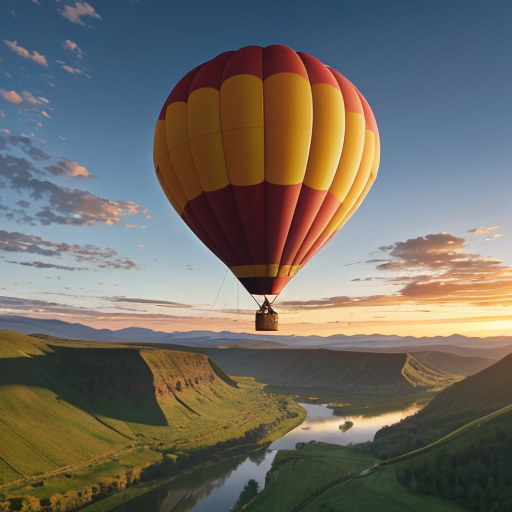}} \\
        \midrule
        \begin{minipage}{1\linewidth}
            \textit{Generate an image of a modern city skyline at night with illuminated skyscrapers. The result should be a high-quality image.}
        \end{minipage} & \raisebox{-0.5\height}{N/A} & \raisebox{-0.5\height}{\includegraphics[width=0.15\linewidth]{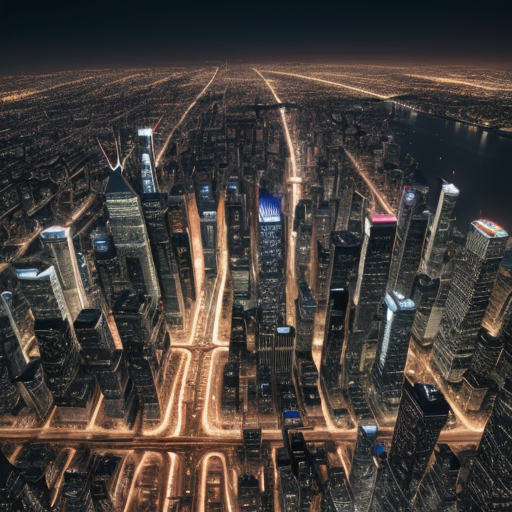}} \\
        \midrule
        \begin{minipage}{1\linewidth}
            \textit{You are given an image of a scribble flower. Repaint the scribble into a realistic red flower. The result should be an image of a red flower.}
        \end{minipage} & \raisebox{-0.5\height}{\includegraphics[width=0.15\linewidth]{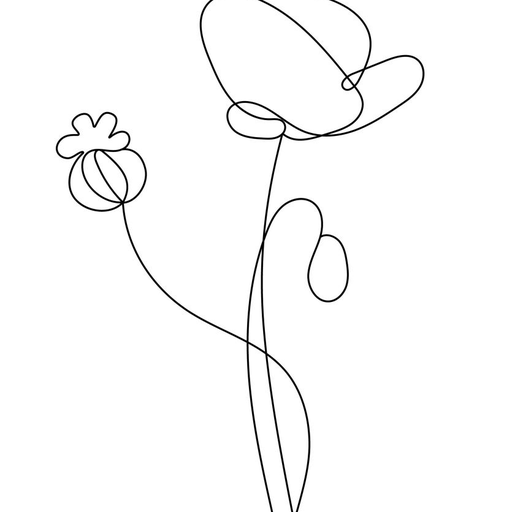}} & \raisebox{-0.5\height}{\includegraphics[width=0.15\linewidth]{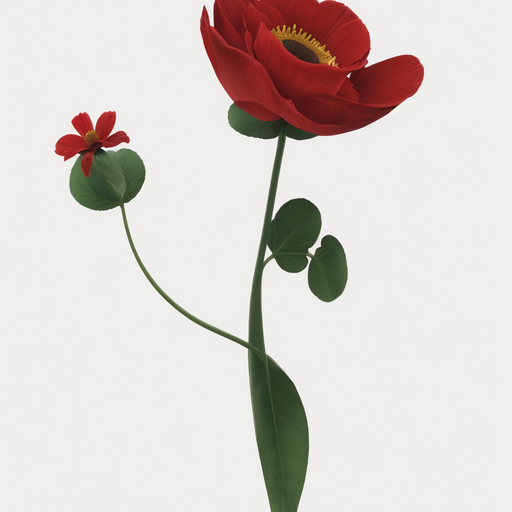}} \\
        \midrule
        \begin{minipage}{1\linewidth}
            \textit{You are given an image of a red apple. Change it into a green apple on a table while maintaining other details. The result should be an image of a green apple.}
        \end{minipage} & \raisebox{-0.5\height}{\includegraphics[width=0.15\linewidth]{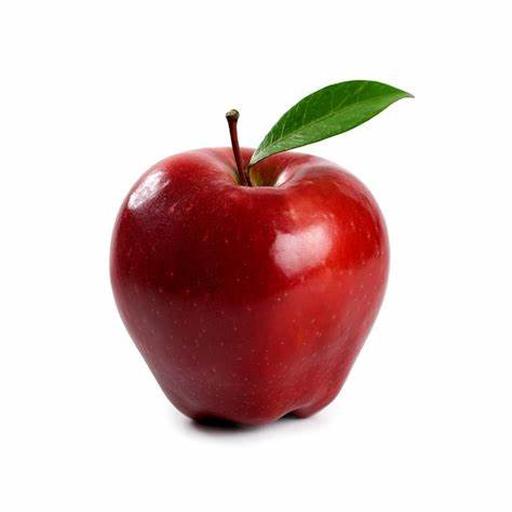}} & \raisebox{-0.5\height}{\includegraphics[width=0.15\linewidth]{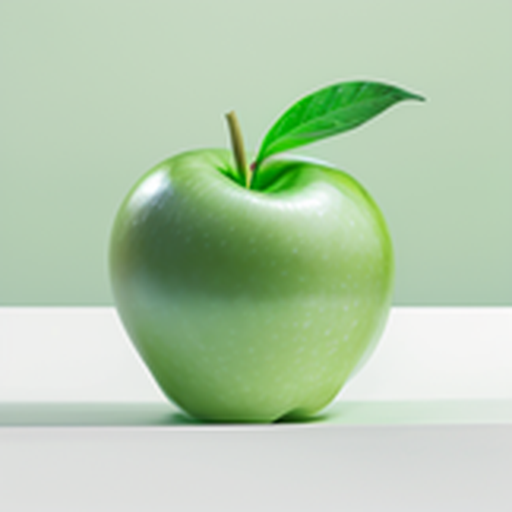}} \\
        \midrule
        \begin{minipage}{1\linewidth}
            \textit{You are given an image of a sample logo containing a bird pattern. Convert it into a cubist art poster with dark colors. The result should be an image of a poster without watermark.}
        \end{minipage} & \raisebox{-0.5\height}{\includegraphics[width=0.15\linewidth]{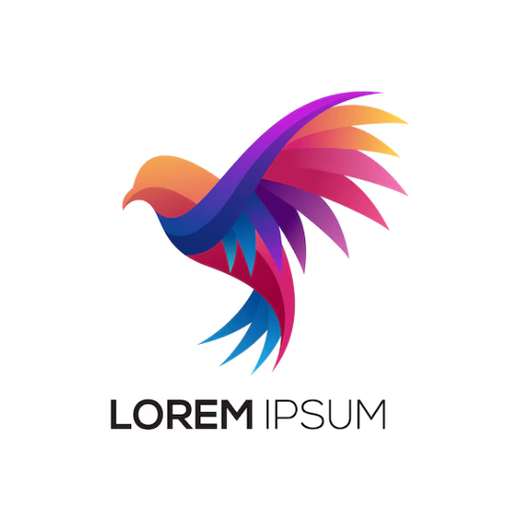}} & \raisebox{-0.5\height}{\includegraphics[width=0.15\linewidth]{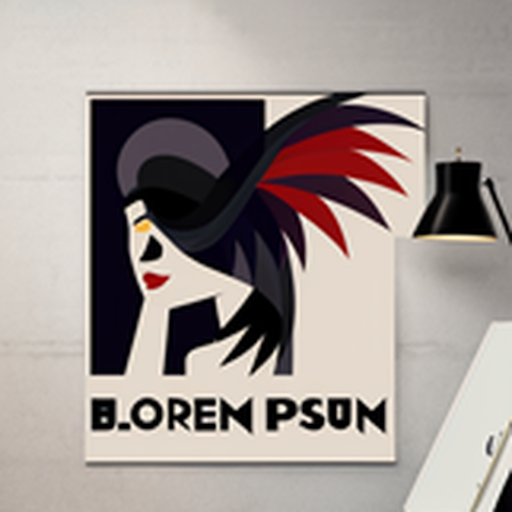}} \\
        \midrule
        \begin{minipage}{1\linewidth}
            \textit{You are given an image of a large castle standing on top of a hill. Convert the castle into the style of ice cream while maintaining its original structure. The result should be an image with the castle transformed into a colorful and fantastic ice cream castle.}
        \end{minipage} & \raisebox{-0.5\height}{\includegraphics[width=0.15\linewidth]{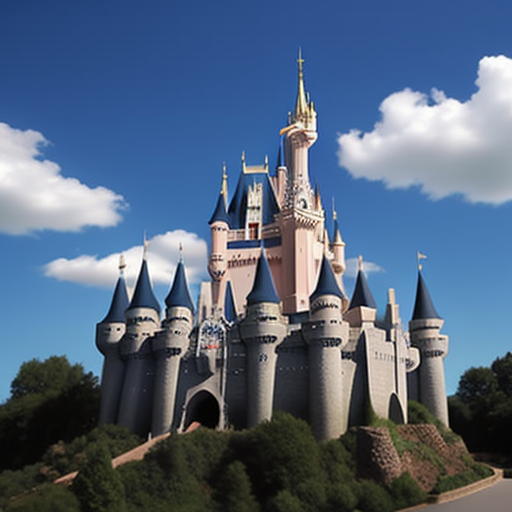}} & \raisebox{-0.5\height}{\includegraphics[width=0.15\linewidth]{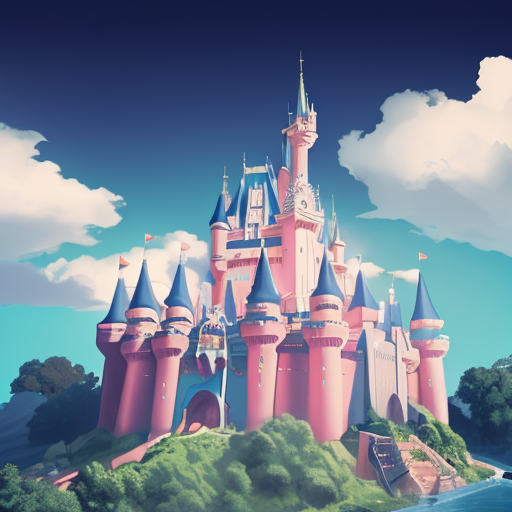}} \\
        \midrule
        \begin{minipage}{1\linewidth}
            \textit{You are given a low-resolution photo of a crowd of people. Upscale the image by 4x. The result should be a high-resolution version of the image.}
        \end{minipage} & \raisebox{-0.5\height}{\includegraphics[width=0.15\linewidth]{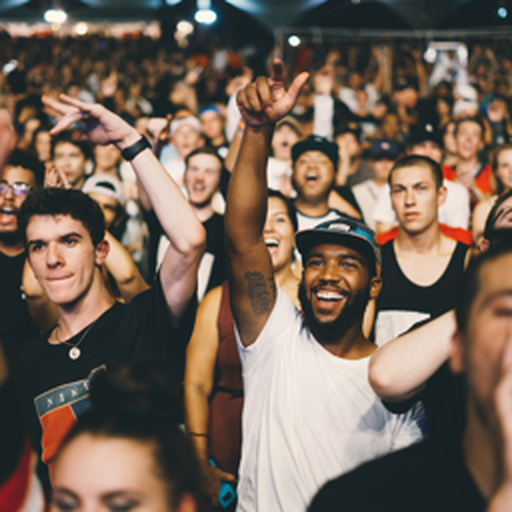}} & \raisebox{-0.5\height}{\includegraphics[width=0.15\linewidth]{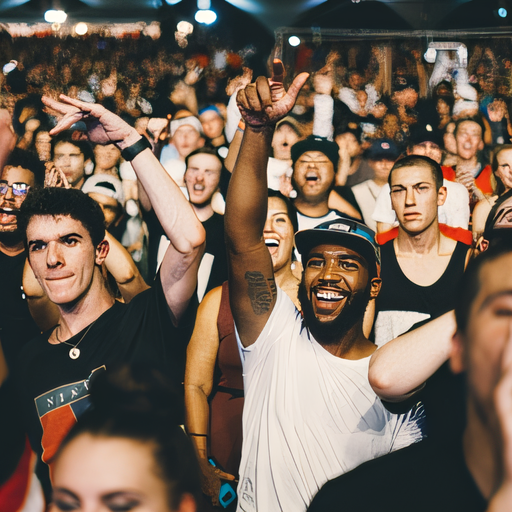}} \\
        \bottomrule
    \end{tabular}
\end{table*}

\clearpage

\begin{table*}
    \caption*{Table~\ref*{table:more_examples}. Continued from previous page.}
    \centering
    \begin{tabular}{p{0.6\linewidth}cc}
        \toprule
        \textbf{Task Instruction} & \textbf{Input} & \textbf{Result} \\
        \midrule
        \begin{minipage}{1\linewidth}
            \textit{You are given an image of a table filled with dishes. Remove the fork on the table. The result should be a high-quality image without visible artifacts.}
        \end{minipage} & \raisebox{-0.5\height}{\includegraphics[width=0.15\linewidth]{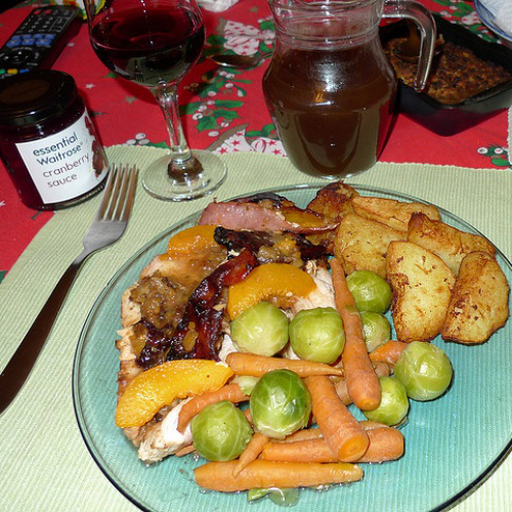}} & \raisebox{-0.5\height}{\includegraphics[width=0.15\linewidth]{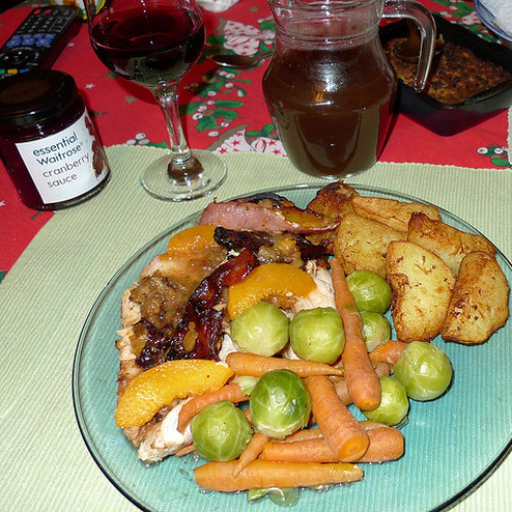}} \\
        \midrule
        \begin{minipage}{1\linewidth}
            \textit{You are given an image of a red car parked on the street. Replace the tree behind the car with a white house. The result should be a high-quality image without visible artifacts.}
        \end{minipage} & \raisebox{-0.5\height}{\includegraphics[width=0.15\linewidth]{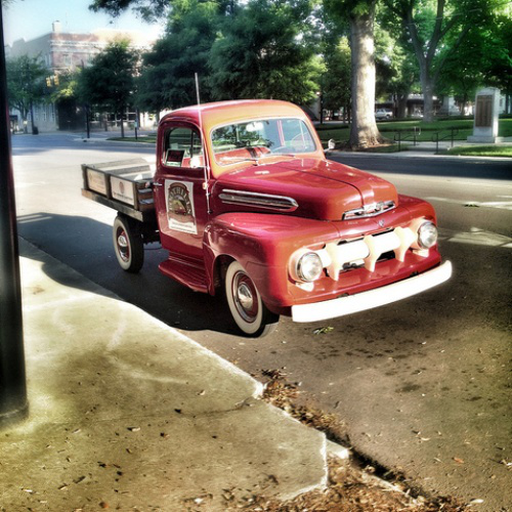}} & \raisebox{-0.5\height}{\includegraphics[width=0.15\linewidth]{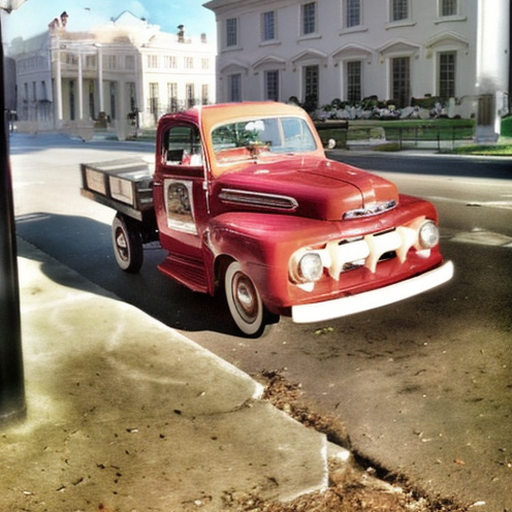}} \\
        \midrule
        \begin{minipage}{1\linewidth}
            \textit{You are given a photo of mountains and rivers with a visible watermark in the bottom right corner. Remove the watermark from the image while maintaining the quality and content of the original photo. The result should be a high-quality image without the watermark.}
        \end{minipage} & \raisebox{-0.5\height}{\includegraphics[width=0.15\linewidth]{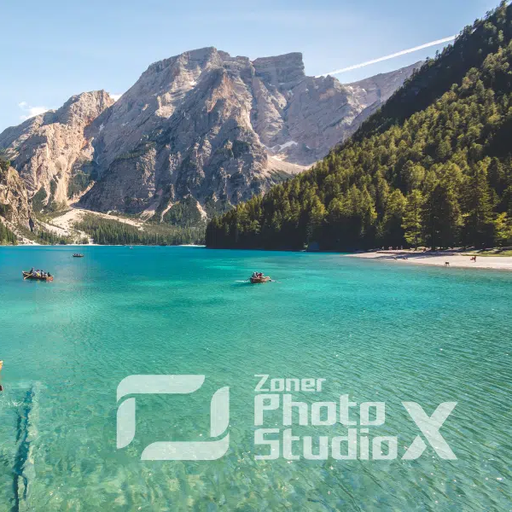}} & \raisebox{-0.5\height}{\includegraphics[width=0.15\linewidth]{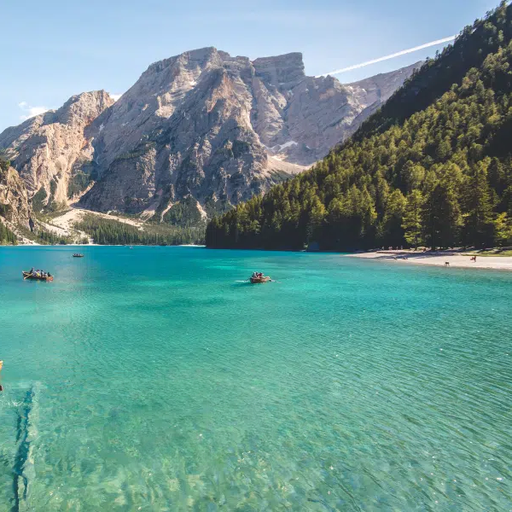}} \\
        \midrule
        \begin{minipage}{1\linewidth}
            \textit{You are given an image of a girl playing the guitar. Generate an image of an old man playing the guitar in a forest with the same pose as the girl. The result should be a realistic image of an old man playing the guitar.}
        \end{minipage} & \raisebox{-0.5\height}{\includegraphics[width=0.15\linewidth]{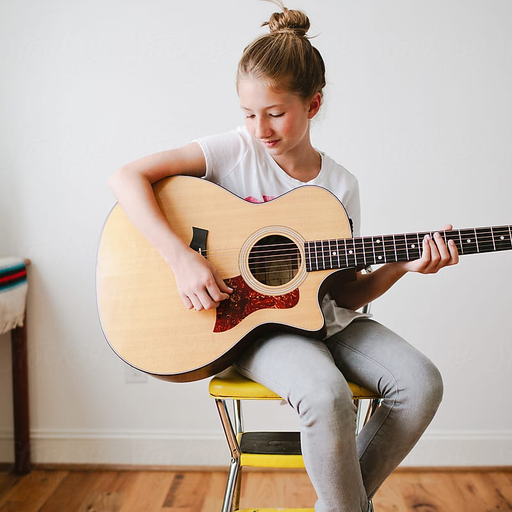}} & \raisebox{-0.5\height}{\includegraphics[width=0.15\linewidth]{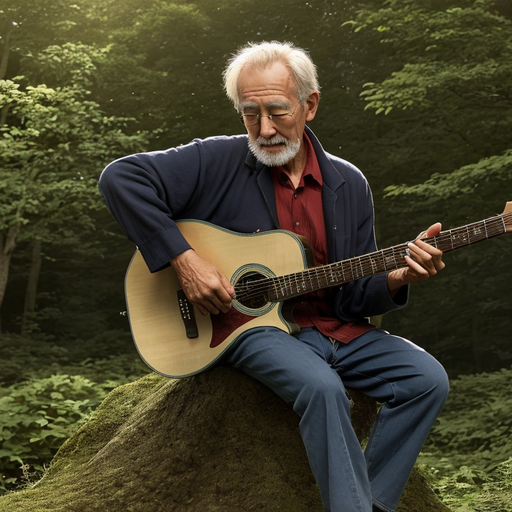}} \\
        \midrule
        \begin{minipage}{1\linewidth}
            \textit{You are given an image of a man wearing a black jacket. Change the black jacket into a white hoodie while ensuring that the modification looks natural and realistic. The result should be a high-quality image of the man wearing a white hoodie.}
        \end{minipage} & \raisebox{-0.5\height}{\includegraphics[width=0.15\linewidth]{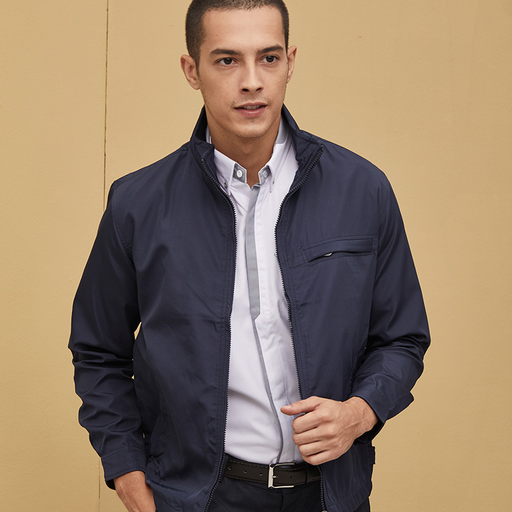}} & \raisebox{-0.5\height}{\includegraphics[width=0.15\linewidth]{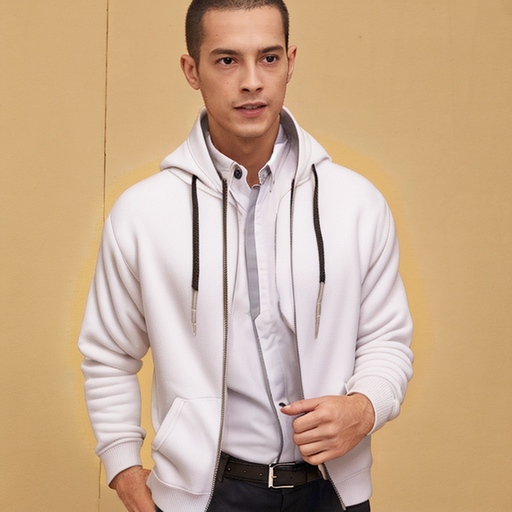}} \\
        \midrule
        \begin{minipage}{1\linewidth}
            \textit{You are given an image of a toy dog. Replace the background with a scene of a sunny park with green grass while keeping the lighting and shadows consistent. The result should be an image of the toy dog in the park scene.}
        \end{minipage} & \raisebox{-0.5\height}{\includegraphics[width=0.15\linewidth]{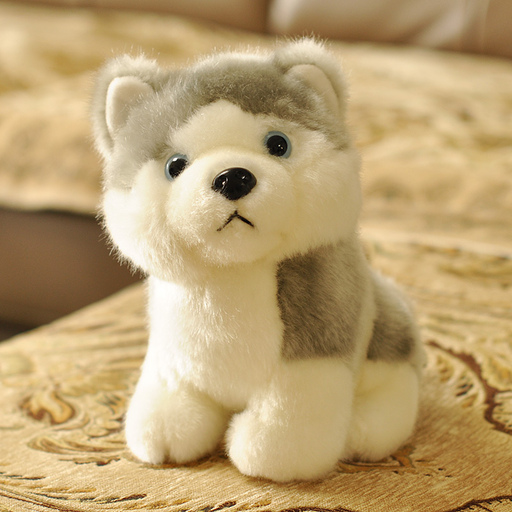}} & \raisebox{-0.5\height}{\includegraphics[width=0.15\linewidth]{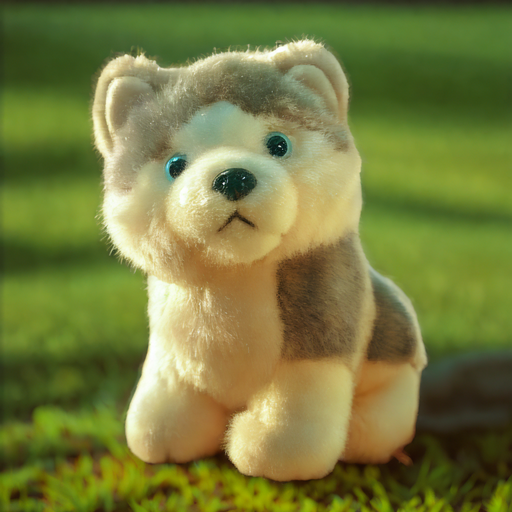}} \\
        \midrule
        \begin{minipage}{1\linewidth}
            \textit{You are given an image containing two bottles of cosmetic products illuminated by a soft yellow light. Modify the illumination into a bright pink light to create a more vibrant and attractive appearance. The result should be an image of the cosmetic products with the new illumination.}
        \end{minipage} & \raisebox{-0.5\height}{\includegraphics[width=0.15\linewidth]{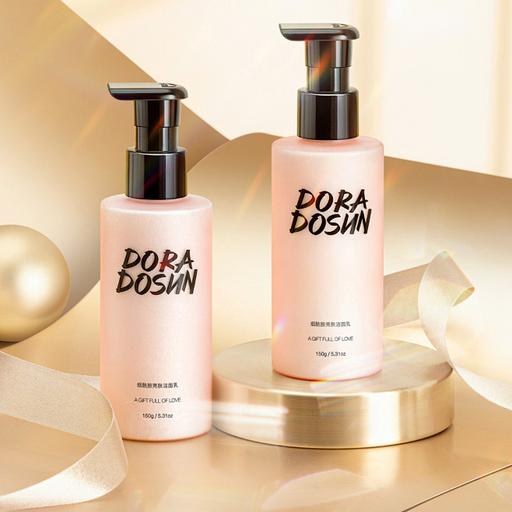}} & \raisebox{-0.5\height}{\includegraphics[width=0.15\linewidth]{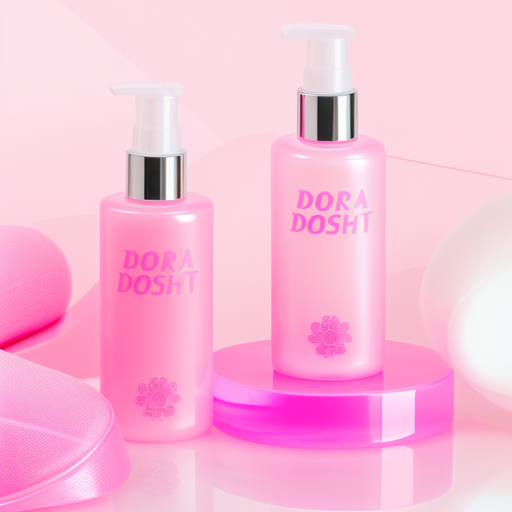}} \\
        \bottomrule
    \end{tabular}
\end{table*}

\clearpage

\begin{tcolorbox}[title=Evaluation Prompt, breakable, colback=bluebg, colbacktitle=bluebg, colframe=bluefg, fonttitle=\bfseries\color{black}, float*=ht, width=\textwidth]
\begin{Verbatim}[breaklines=true, breakanywhere=true, fontsize=\footnotesize]
You are an expert in image and video generation, familiar with the latest tasks and techniques. You are capable of understanding the task instruction, analyzing the generation result, and providing an accurate evaluation. Now you are evaluating the result of a text-to-image generation task. You should be tolerant to the quality of the generation result, and focus on the consistency with the instruction.

The task instruction is described as: First generate an image of a ship floating in the sea. The ship should be placed in the right half of the image, with the sea filling other areas. Then convert the image into a painting with watercolor style. The result should be a high-quality image of the ship as a watercolor painting.

The given image is the generation result, with an actual resolution of 512x512.

First, analyze whether the generation result meets each key point in the instruction. Enclose your analysis in the <analysis> tag. For example: <analysis>There is a cat in an astronaut suit, which is consistent with the instruction. The wall is white, which is different from the "green wall" in the instruction.</analysis>.

Then, provide a final judgment of whether the generation result complies with the instruction. The judgment should either be "True" or "False". Enclose your judgment in the <judgment> tag. For example: <judgment>False</judgment>.
\end{Verbatim}
\begin{center}
\includegraphics[width=0.48\linewidth]{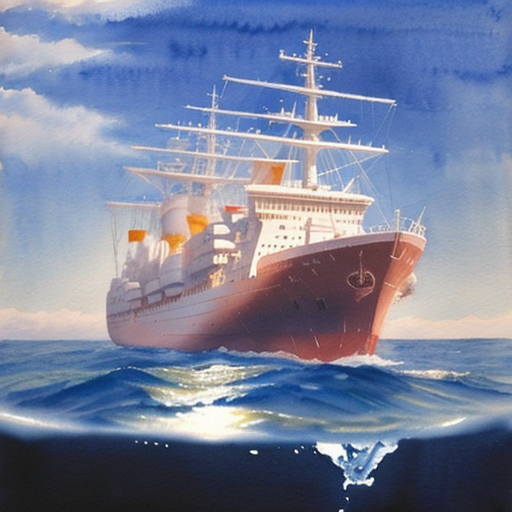}
\end{center}
\end{tcolorbox}

\begin{tcolorbox}[title=GPT-4o Answer, breakable, colback=redbg, colbacktitle=redbg, colframe=redfg, fonttitle=\bfseries\color{black}, float*=ht, width=\textwidth]
\begin{Verbatim}[breaklines=true, breakanywhere=true, fontsize=\footnotesize]
<analysis>There is a ship floating in the sea, which matches the first part of the instruction. However, the ship is positioned in the center of the image, not on the right half as specified in the instruction. The sea fills a significant part of the image underneath the ship but does not fill the entire left half as required. The painting is shown in a watercolor style, which satisfies the final part of the instruction. Overall, the main issue is the placement of the ship, which should have been shifted to the right side of the image.</analysis>

<judgment>False</judgment>
\end{Verbatim}
\captionsetup[figure]{hypcap=false}
\captionof{figure}{A typical example for VLM-based evaluation in ComfyBench.}
\label{figure:vlmbased_evaluation}
\end{tcolorbox}

\clearpage

\begin{tcolorbox}[title=PlanAgent Prompt, breakable, float*=ht, width=\textwidth]
\begin{Verbatim}[breaklines=true, breakanywhere=true, fontsize=\footnotesize]
## Task

ComfyUI uses workflows to create and execute Stable Diffusion pipelines so that users can design their own workflows to generate highly customized artworks. ComfyUI workflows can be formulated into the equivalent Python code, where each statement represents the execution of a single node. You are an expert in ComfyUI, helping users to design workflows according to their requirements.

Now you are required to create a ComfyUI workflow to finish the following task:

{instruction}

The core requirements and the expected paradigm are analyzed as follows:

{analysis}

Improving the quality of the generation result with additional steps, such as upscaling and interpolation, is not recommended, unless specified in the requirements.

## Reference

According to the requirements, we have retrieved some relevant workflows which may be helpful:

{reference}

## History

Here is a recent history of your thought, plan and action in the previous steps. The most recent record is at the bottom.

{history}

## Workspace

The code and annotation of the current workflow you are working on are presented as follows:

{workspace}

## Action

Based on the history and workspace, you should first think about what functions have been implemented and what modules remain to be added. Your thought should be enclosed with "<thought>" tag. For example: <thought>The basic pipeline has been implemented, but a module is needed to improve the quality.</thought>.

After that, you should update your step-by-step plan to further modify your workflow. There are {limitation} steps remaining, so your plan should contain at most {limitation} steps. Make sure that each step is feasible to be converted into a single action. Your plan should be enclosed with "<plan>" tag. For example: <plan>Step 1: I will refer to "reference_name" to add a module. Step 2: I will finish the task since the expected effects are realized.</plan>.

Finally, you should choose one of the following actions and specify the arguments (if required), so that the updated workflow can realize the first step in your plan. You should provide your action with the format of function calls in Python. Your action should be enclosed with "<action>" tag. For example: <action>combine(name="reference_name")</action>, <action>adapt(prompt="Change the factor to 0.5 and rewrite the prompt.")</action>, and <action>finish()</action>.

- `load`: Load a reference workflow into the workspace to replace the current workflow, so that you can start over. Arguments:
  - `name`: The name of the reference workflow you want to load.
- `combine`: Combine the current workflow with a reference workflow, so that necessary modules can be added. Arguments:
  - `name`: The name of the reference workflow you want to combine.
- `adapt`: Adapt some parameters in the current workflow, so that the expected effects can be realized. Arguments:
  - `prompt`: The prompt to specify the adaptation you want to make.
- `retrieve`: Retrieve a new batch of reference workflows, so that more useful references can be found. Arguments:
  - `prompt`: The prompt to describe the reference you want to retrieve.
- `finish`: Finish the task since the current workflow can realize the expected effects.

Refer to the history before making a decision. Here are some general rules you should follow:

1. You should choose the `load` action if and only if the history is empty.
2. If you choose the `load` or `combine` action, make sure the name exists in the reference. Otherwise, try to update the reference with the `retrieve` action.
3. You should not choose the `adapt` action twice in a row, because they can be simplified into a single action.
4. If you choose the `adapt` or `retrieve` action, make sure the prompt is concise and contains all the necessary information.
5. You should choose the `finish` action before the remaining steps count down to 0.

Now, provide your thought, plan and action with the required format.
\end{Verbatim}
\captionsetup[figure]{hypcap=false}
\captionof{figure}{Prompt template for PlanAgent.}
\label{figure:planagent_prompt}
\end{tcolorbox}

\clearpage

\begin{tcolorbox}[title=CombineAgent Prompt, breakable, float*=ht, width=\textwidth]
\begin{Verbatim}[breaklines=true, breakanywhere=true, fontsize=\footnotesize]
## Task

ComfyUI uses workflows to create and execute Stable Diffusion pipelines so that users can design their own workflows to generate highly customized artworks. ComfyUI workflows can be formulated into the equivalent Python code, where each statement represents the execution of a single node. You are an expert in ComfyUI, helping users to design workflows according to their requirements.

Now you are required to create a ComfyUI workflow to finish the following task:

{instruction}

The core requirements and the expected paradigm are analyzed as follows:

{analysis}

## Reference

The code and annotation of the current workflow you are referring to are presented as follows:

{reference}

## Workspace

The code and annotation of the current workflow you are working on are presented as follows:

{workspace}

## Combination

Based on the current working progress, your schedule is presented as follows:

{schedule}

You are working on the first step of your schedule. In other words, you should combine the reference workflow with the current workflow according to your schedule.

First, you should provide your Python code to formulate the updated workflow. Each line of code should correspond to a single node, so you should avoid nested calls in a single statement. You should also avoid reusing the same variable name, even if the variable is temporary. Your code should be enclosed with "<code>" tag. For example: <code>output = node(input)</code>.

After that, you should provide an annotation as in the reference, including the function and principle of the updated workflow. The function should be enclosed with "<function>" tag. For example: <function>This workflow generates a high-resolution image of a running horse.</function>. The principle should be enclosed with "<principle>" tag. For example: <principle>The workflow first generates a low-resolution image using the text-to-image pipeline and then applies an upscaling module to improve the resolution.</principle>.

Now, provide your code and annotation with the required format.
\end{Verbatim}
\captionsetup[figure]{hypcap=false}
\captionof{figure}{Prompt template for CombineAgent.}
\label{figure:combineagent_prompt}
\end{tcolorbox}

\clearpage

\begin{tcolorbox}[title=AdaptAgent Prompt, breakable, float*=ht, width=\textwidth]
\begin{Verbatim}[breaklines=true, breakanywhere=true, fontsize=\footnotesize]
## Task

ComfyUI uses workflows to create and execute Stable Diffusion pipelines so that users can design their own workflows to generate highly customized artworks. ComfyUI workflows can be formulated into the equivalent Python code, where each statement represents the execution of a single node. You are an expert in ComfyUI, helping users to design workflows according to their requirements.

Now you are required to create a ComfyUI workflow to finish the following task:

{instruction}

The core requirements and the expected paradigm are analyzed as follows:

{analysis}

## Workspace

The code and annotation of the current workflow you are working on are presented as follows:

{workspace}

## Adaptation

Based on the current working progress, your schedule is presented as follows:

{schedule}

You are working on the first step of your schedule. In other words, you should modify the parameters in the current workflow according to your schedule. The adaptation you want to make is specified as follows:

{adaptation}

First, you should provide your Python code to formulate the updated workflow. Each line of code should correspond to a single node, so you should avoid nested calls in a single statement. You should also avoid reusing the same variable name, even if the variable is temporary. Your code should be enclosed with "<code>" tag. For example: <code>output = node(input)</code>.

After that, you should provide an annotation as in the reference, including the function and principle of the updated workflow. The function should be enclosed with "<function>" tag. For example: <function>This workflow generates a high-resolution image of a running horse.</function>. The principle should be enclosed with "<principle>" tag. For example: <principle>The workflow first generates a low-resolution image using the text-to-image pipeline and then applies an upscaling module to improve the resolution.</principle>.

Now, provide your code and annotation with the required format.
\end{Verbatim}
\captionsetup[figure]{hypcap=false}
\captionof{figure}{Prompt template for AdaptAgent.}
\label{figure:adaptagent_prompt}
\end{tcolorbox}

\clearpage

\begin{tcolorbox}[title=RefineAgent Prompt, breakable, float*=ht, width=\textwidth]
\begin{Verbatim}[breaklines=true, breakanywhere=true, fontsize=\footnotesize]
## Task

ComfyUI uses workflows to create and execute Stable Diffusion pipelines so that users can design their own workflows to generate highly customized artworks. ComfyUI workflows can be formulated into the equivalent Python code, where each statement represents the execution of a single node. You are an expert in ComfyUI, helping users to design workflows according to their requirements.

Now you are required to create a ComfyUI workflow to finish the following task:

{instruction}

The core requirements and the expected paradigm are analyzed as follows:

{analysis}

## Reference

According to the requirements, we have retrieved some relevant workflows which may be helpful:

{reference}

## Workspace

The code and annotation of the current workflow you are working on are presented as follows:

{workspace}

## Refinement

An error is detected in the current workflow, which is caused by some bugs in the Python code, such as nested calls and missing parameters. The specific error message is presented as follows:

{refinement}

First, you should explain the reason of the error. Your explanation should be enclosed with "<explanation>" tag. For example: <explanation>The error is caused by the missing input.</explanation>.

After that, you should provide the corrected Python code to formulate the updated workflow. Each line of code should correspond to a single node, so you should avoid nested calls in a single statement. You should also avoid reusing the same variable name, even if the variable is temporary. Your code should be enclosed with "<code>" tag. For example: <code>output = node(input)</code>.

Finally, you should provide an annotation as in the reference, including the function and principle of the updated workflow. The function should be enclosed with "<function>" tag. For example: <function>This workflow generates a high-resolution image of a running horse.</function>. The principle should be enclosed with "<principle>" tag. For example: <principle>The workflow first generates a low-resolution image using the text-to-image pipeline and then applies an upscaling module to improve the resolution.</principle>.

Now, provide your explanation, code, and annotation with the required format.
\end{Verbatim}
\captionsetup[figure]{hypcap=false}
\captionof{figure}{Prompt template for RefineAgent.}
\label{figure:refineagent_prompt}
\end{tcolorbox}

\clearpage

\begin{tcolorbox}[title=Task Instruction, breakable, colback=yellowbg, colbacktitle=yellowbg, colframe=yellowfg, fonttitle=\bfseries\color{black}, float*=ht, width=\textwidth]
\begin{Verbatim}[breaklines=true, breakanywhere=true, fontsize=\footnotesize]
First generate a 2-second video of colorful fish swimming in an aquarium with bubbles rising to the surface. Then interpolate the video to increase the frame rate by 3x. The result should be a smoother video of the swimming fish.
\end{Verbatim}
\end{tcolorbox}

\begin{tcolorbox}[title=PlanAgent Prompt, breakable, colback=bluebg, colbacktitle=bluebg, colframe=bluefg, fonttitle=\bfseries\color{black}, float*=ht, width=\textwidth]
\begin{Verbatim}[breaklines=true, breakanywhere=true, fontsize=\footnotesize]
## Task

ComfyUI uses workflows to create and execute Stable Diffusion pipelines so that users can design their own workflows to generate highly customized artworks. ComfyUI workflows can be formulated into the equivalent Python code, where each statement represents the execution of a single node. You are an expert in ComfyUI, helping users to design workflows according to their requirements.

Now you are required to create a ComfyUI workflow to finish the following task:

First generate a 2-second video of colorful fish swimming in an aquarium with bubbles rising to the surface. Then interpolate the video to increase the frame rate by 3x. The result should be a smoother video of the swimming fish.

The core requirements and the expected paradigm are analyzed as follows:

The task requires generating a short 2-second video of fish swimming in an aquarium with bubbles rising using a text-to-video approach, as the primary input is textual. The video needs to depict dynamic, colorful fish movements with the additional visual element of rising bubbles. After generating this video, the frame rate needs to be increased by 3x using a video interpolation process to achieve smoother motion. Thus, the workflow core involves two stages, first, synthesizing the video from a text prompt using a text-to-video node or series of nodes, followed by applying video interpolation to enhance frame smoothness. There are no specific quality-oriented steps like upscaling or refinement beyond the interpolation requirement.

Improving the quality of the generation result with additional steps, such as upscaling and interpolation, is not recommended, unless specified in the requirements.

## Reference

According to the requirements, we have retrieved some relevant workflows which may be helpful:

- Example: text_to_video

<function>
This workflow follows a text-to-video paradigm, where it generates a video from a given text prompt. It first creates an initial image based on the text description and then uses that image as the starting frame to generate a video with motion and transformation using a Stable Video Diffusion model. The output is a 3-second video at 8 frames per second, depicting beautiful scenery with mountains, rivers, and clouds.
</function>

<principle>
The workflow first loads two models: a Stable Diffusion model for generating the initial image from the text ("sd_xl_base_1.0.safetensors") and a Stable Video Diffusion model ("svd_xt_1_1.safetensors") for video generation. It uses the text description to create conditioning, generating an initial 1024x576 image. The image is decoded from latent space via a VAE. The video-specific node ("SVD_img2vid_Conditioning") then applies continuity and motion to this image, producing conditioned latent representations for video generation. The final video is created by sampling the latent space over multiple frames and combining them into an MP4 video using the specified frame rate and format.
</principle>

- Example: video_frame_interpolation
\end{Verbatim}
\end{tcolorbox}

\clearpage

\begin{tcolorbox}[breakable, colback=bluebg, colbacktitle=bluebg, colframe=bluefg, fonttitle=\bfseries\color{black}, float*=ht, width=\textwidth]
\begin{Verbatim}[breaklines=true, breakanywhere=true, fontsize=\footnotesize]

<function>
This workflow performs video frame interpolation using the RIFE VFI model. It takes an input video such as "play_guitar.gif", increases the frame rate by generating intermediate frames (interpolating) with a multiplier (in this case, 3x), and produces a smoother video with a higher frame rate (from 8 to 24 frames per second). The final output is saved as a new video or animated GIF.
</function>

<principle>
The workflow first loads the input video using "VHS_LoadVideo", which extracts the individual frames. The "RIFE VFI" node is then used to interpolate the frames by generating additional frames between the existing ones. In this scenario, the multiplier is set to 3x, effectively tripling the frame count and enabling a smoother video playback at 24 frames per second. Finally, the interpolated frames are combined into a video or GIF format using "VHS_VideoCombine".
</principle>

- Example: image_to_video

<function>
This workflow follows an image-to-video paradigm. It requires an input image (in this case, "play_guitar.jpg") and generates a 4-second video at 6 frames per second (24 video frames in total) based on that image. The workflow outputs the generated video.
</function>

<principle>
The workflow uses the "svd_xt_1_1.safetensors" Stable Video Diffusion model to generate a video from the input image "play_guitar.jpg". The "SVD_img2vid_Conditioning" node creates the necessary conditioning for video generation, including the number of frames, resolution, and motion characteristics. A KSamplerAdvanced node adds noise and performs generative sampling over multiple steps to create diverse video frames. These frames are then decoded back into images via a VAE, and finally, the "VHS_VideoCombine" node compiles these images into a 4-second video at 8 frames per second.
</principle>

- Example: text_to_image

<function>
This workflow implements a basic text-to-image generation pipeline using Stable Diffusion. It requires both positive (desired traits) and negative (undesired traits) text prompts to generate an image. In this specific case, the workflow will output a high-resolution photo of a cat wearing a spacesuit inside a spaceship, avoiding blurry or illustration-like effects.
</function>

<principle>
The workflow begins by loading the "dreamshaper_8.safetensors" Stable Diffusion model. It generates a blank latent space as the starting point for the image generation. The positive prompt ("a photo of a cat wearing a spacesuit inside a spaceship") and negative prompt ("blurry, illustration") are encoded into conditioning by the CLIPTextEncode node. The KSampler node then uses these conditionings to guide the generative process over 20 sampling steps, applying denoising to the latent space. The resulting latent code is subsequently decoded into an image using the VAE and saved to disk.
</principle>

- Example: scribble_image_repaint

<function>
This workflow follows a sketch-to-image paradigm, where it takes a scribble image ("simple_graffiti.png") along with a text prompt ("a bird, open wings") and generates a detailed, high-quality image based on both the scribble and the prompt.
</function>

<principle>
\end{Verbatim}
\end{tcolorbox}

\clearpage

\begin{tcolorbox}[breakable, colback=bluebg, colbacktitle=bluebg, colframe=bluefg, fonttitle=\bfseries\color{black}, float*=ht, width=\textwidth]
\begin{Verbatim}[breaklines=true, breakanywhere=true, fontsize=\footnotesize]
The workflow first loads and inverts the input scribble image. The inverted image is then used by a "control_v11p_sd15_scribble_fp16" ControlNet model to extract and applied to the scribble, controlling and guiding the image generation process. A pre-trained model ("dreamshaper_8.safetensors") processes the positive conditioning generated by the combination of the text prompt and the ControlNet's output, along with negative conditioning from undesired traits such as "horror" and "lowres." The latent space image is generated using a KSampler and further decoded by the VAE to form the final image. The model synthesizes a highly detailed image while adhering closely to both the shape of the scribble and the desired features implied by the text prompt.
</principle>

## History

Here is a recent history of your thought, plan and action in the previous steps. The most recent record is at the bottom.

- The history is empty.

## Workspace

The code and annotation of the current workflow you are working on are presented as follows:

- The workspace is empty.

## Action

Based on the history and workspace, you should first think about what functions have been implemented and what modules remain to be added. Your thought should be enclosed with "<thought>" tag. For example: <thought>The basic pipeline has been implemented, but a module is needed to improve the quality.</thought>.

After that, you should update your step-by-step plan to further modify your workflow. There are 5 steps remaining, so your plan should contain at most 5 steps. Make sure that each step is feasible to be converted into a single action. Your plan should be enclosed with "<plan>" tag. For example: <plan>Step 1: I will refer to "reference_name" to add a module. Step 2: I will finish the task since the expected effects are realized.</plan>.

Finally, you should choose one of the following actions and specify the arguments (if required), so that the updated workflow can realize the first step in your plan. You should provide your action with the format of function calls in Python. Your action should be enclosed with "<action>" tag. For example: <action>combine(name="reference_name")</action>, <action>adapt(prompt="Change the factor to 0.5 and rewrite the prompt.")</action>, and <action>finish()</action>.

- `load`: Load a reference workflow into the workspace to replace the current workflow, so that you can start over. Arguments:
  - `name`: The name of the reference workflow you want to load.
- `combine`: Combine the current workflow with a reference workflow, so that necessary modules can be added. Arguments:
  - `name`: The name of the reference workflow you want to combine.
- `adapt`: Adapt some parameters in the current workflow, so that the expected effects can be realized. Arguments:
  - `prompt`: The prompt to specify the adaptation you want to make.
- `retrieve`: Retrieve a new batch of reference workflows, so that more useful references can be found. Arguments:
  - `prompt`: The prompt to describe the reference you want to retrieve.
- `finish`: Finish the task since the current workflow can realize the expected effects.

Refer to the history before making a decision. Here are some general rules you should follow:

1. You should choose the `load` action if and only if the history is empty.
2. If you choose the `load` or `combine` action, make sure the name exists in the reference. Otherwise, try to update the reference with the `retrieve` action.
3. You should not choose the `adapt` action twice in a row, because they can be simplified into a single action.
\end{Verbatim}
\end{tcolorbox}

\clearpage

\begin{tcolorbox}[breakable, colback=bluebg, colbacktitle=bluebg, colframe=bluefg, fonttitle=\bfseries\color{black}, float*=ht, width=\textwidth]
\begin{Verbatim}[breaklines=true, breakanywhere=true, fontsize=\footnotesize]
4. If you choose the `adapt` or `retrieve` action, make sure the prompt is concise and contains all the necessary information.
5. You should choose the `finish` action before the remaining steps count down to 0.

Now, provide your thought, plan and action with the required format.
\end{Verbatim}
\end{tcolorbox}

\begin{tcolorbox}[title=PlanAgent Answer, breakable, colback=redbg, colbacktitle=redbg, colframe=redfg, fonttitle=\bfseries\color{black}, float*=ht, width=\textwidth]
\begin{Verbatim}[breaklines=true, breakanywhere=true, fontsize=\footnotesize]
<thought>Since the history is empty, I need to start by loading a relevant reference workflow that can be used as a base for generating the 2-second video of colorful fish in an aquarium. The "text_to_video" reference seems appropriate to use as it involves generating video from text prompts, which aligns with the first part of the task.</thought>

<plan>Step 1: I will load the "text_to_video" reference workflow to use it as the base.  
Step 2: I will adapt the workflow to modify the prompt to match the requirements ("colorful fish swimming in an aquarium with bubbles rising").  
Step 3: I will combine it with the "video_frame_interpolation" reference to enable interpolation for smoother video.  
Step 4: I will adapt the interpolation multiplier to 3x to achieve the expected frame rate increase.  
Step 5: I will finish the task by confirming the generated output meets the required specifications.</plan>

<action>load(name="text_to_video")</action>
\end{Verbatim}
\end{tcolorbox}

\begin{tcolorbox}[title=PlanAgent Prompt, breakable, colback=bluebg, colbacktitle=bluebg, colframe=bluefg, fonttitle=\bfseries\color{black}, float*=ht, width=\textwidth]
\begin{Verbatim}[breaklines=true, breakanywhere=true, fontsize=\footnotesize]
## Task

ComfyUI uses workflows to create and execute Stable Diffusion pipelines so that users can design their own workflows to generate highly customized artworks. ComfyUI workflows can be formulated into the equivalent Python code, where each statement represents the execution of a single node. You are an expert in ComfyUI, helping users to design workflows according to their requirements.

Now you are required to create a ComfyUI workflow to finish the following task:

First generate a 2-second video of colorful fish swimming in an aquarium with bubbles rising to the surface. Then interpolate the video to increase the frame rate by 3x. The result should be a smoother video of the swimming fish.

The core requirements and the expected paradigm are analyzed as follows:

The task requires generating a short 2-second video of fish swimming in an aquarium with bubbles rising using a text-to-video approach, as the primary input is textual. The video needs to depict dynamic, colorful fish movements with the additional visual element of rising bubbles. After generating this video, the frame rate needs to be increased by 3x using a video interpolation process to achieve smoother motion. Thus, the workflow core involves two stages, first, synthesizing the video from a text prompt using a text-to-video node or series of nodes, followed by applying video interpolation to enhance frame smoothness. There are no specific quality-oriented steps like upscaling or refinement beyond the interpolation requirement.

Improving the quality of the generation result with additional steps, such as upscaling and interpolation, is not recommended, unless specified in the requirements.

## Reference

According to the requirements, we have retrieved some relevant workflows which may be helpful:

\end{Verbatim}
\end{tcolorbox}

\clearpage

\begin{tcolorbox}[breakable, colback=bluebg, colbacktitle=bluebg, colframe=bluefg, fonttitle=\bfseries\color{black}, float*=ht, width=\textwidth]
\begin{Verbatim}[breaklines=true, breakanywhere=true, fontsize=\footnotesize]
- Example: text_to_video

<function>
This workflow follows a text-to-video paradigm, where it generates a video from a given text prompt. It first creates an initial image based on the text description and then uses that image as the starting frame to generate a video with motion and transformation using a Stable Video Diffusion model. The output is a 3-second video at 8 frames per second, depicting beautiful scenery with mountains, rivers, and clouds.
</function>

<principle>
The workflow first loads two models: a Stable Diffusion model for generating the initial image from the text ("sd_xl_base_1.0.safetensors") and a Stable Video Diffusion model ("svd_xt_1_1.safetensors") for video generation. It uses the text description to create conditioning, generating an initial 1024x576 image. The image is decoded from latent space via a VAE. The video-specific node ("SVD_img2vid_Conditioning") then applies continuity and motion to this image, producing conditioned latent representations for video generation. The final video is created by sampling the latent space over multiple frames and combining them into an MP4 video using the specified frame rate and format.
</principle>

- Example: video_frame_interpolation

<function>
This workflow performs video frame interpolation using the RIFE VFI model. It takes an input video such as "play_guitar.gif", increases the frame rate by generating intermediate frames (interpolating) with a multiplier (in this case, 3x), and produces a smoother video with a higher frame rate (from 8 to 24 frames per second). The final output is saved as a new video or animated GIF.
</function>

<principle>
The workflow first loads the input video using "VHS_LoadVideo", which extracts the individual frames. The "RIFE VFI" node is then used to interpolate the frames by generating additional frames between the existing ones. In this scenario, the multiplier is set to 3x, effectively tripling the frame count and enabling a smoother video playback at 24 frames per second. Finally, the interpolated frames are combined into a video or GIF format using "VHS_VideoCombine".
</principle>

- Example: image_to_video

<function>
This workflow follows an image-to-video paradigm. It requires an input image (in this case, "play_guitar.jpg") and generates a 4-second video at 6 frames per second (24 video frames in total) based on that image. The workflow outputs the generated video.
</function>

<principle>
The workflow uses the "svd_xt_1_1.safetensors" Stable Video Diffusion model to generate a video from the input image "play_guitar.jpg". The "SVD_img2vid_Conditioning" node creates the necessary conditioning for video generation, including the number of frames, resolution, and motion characteristics. A KSamplerAdvanced node adds noise and performs generative sampling over multiple steps to create diverse video frames. These frames are then decoded back into images via a VAE, and finally, the "VHS_VideoCombine" node compiles these images into a 4-second video at 8 frames per second.
</principle>

- Example: text_to_image

<function>
\end{Verbatim}
\end{tcolorbox}

\clearpage

\begin{tcolorbox}[breakable, colback=bluebg, colbacktitle=bluebg, colframe=bluefg, fonttitle=\bfseries\color{black}, float*=ht, width=\textwidth]
\begin{Verbatim}[breaklines=true, breakanywhere=true, fontsize=\footnotesize]
This workflow implements a basic text-to-image generation pipeline using Stable Diffusion. It requires both positive (desired traits) and negative (undesired traits) text prompts to generate an image. In this specific case, the workflow will output a high-resolution photo of a cat wearing a spacesuit inside a spaceship, avoiding blurry or illustration-like effects.
</function>

<principle>
The workflow begins by loading the "dreamshaper_8.safetensors" Stable Diffusion model. It generates a blank latent space as the starting point for the image generation. The positive prompt ("a photo of a cat wearing a spacesuit inside a spaceship") and negative prompt ("blurry, illustration") are encoded into conditioning by the CLIPTextEncode node. The KSampler node then uses these conditionings to guide the generative process over 20 sampling steps, applying denoising to the latent space. The resulting latent code is subsequently decoded into an image using the VAE and saved to disk.
</principle>

- Example: scribble_image_repaint

<function>
This workflow follows a sketch-to-image paradigm, where it takes a scribble image ("simple_graffiti.png") along with a text prompt ("a bird, open wings") and generates a detailed, high-quality image based on both the scribble and the prompt.
</function>

<principle>
The workflow first loads and inverts the input scribble image. The inverted image is then used by a "control_v11p_sd15_scribble_fp16" ControlNet model to extract and applied to the scribble, controlling and guiding the image generation process. A pre-trained model ("dreamshaper_8.safetensors") processes the positive conditioning generated by the combination of the text prompt and the ControlNet's output, along with negative conditioning from undesired traits such as "horror" and "lowres." The latent space image is generated using a KSampler and further decoded by the VAE to form the final image. The model synthesizes a highly detailed image while adhering closely to both the shape of the scribble and the desired features implied by the text prompt.
</principle>

## History

Here is a recent history of your thought, plan and action in the previous steps. The most recent record is at the bottom.

- Step: 1

<thought>
Since the history is empty, I need to start by loading a relevant reference workflow that can be used as a base for generating the 2-second video of colorful fish in an aquarium. The "text_to_video" reference seems appropriate to use as it involves generating video from text prompts, which aligns with the first part of the task.
</thought>

<plan>
Step 1: I will load the "text_to_video" reference workflow to use it as the base.  
Step 2: I will adapt the workflow to modify the prompt to match the requirements ("colorful fish swimming in an aquarium with bubbles rising").  
Step 3: I will combine it with the "video_frame_interpolation" reference to enable interpolation for smoother video.  
Step 4: I will adapt the interpolation multiplier to 3x to achieve the expected frame rate increase.  
Step 5: I will finish the task by confirming the generated output meets the required specifications.
</plan>

<action>
load(name="text_to_video")
\end{Verbatim}
\end{tcolorbox}

\clearpage

\begin{tcolorbox}[breakable, colback=bluebg, colbacktitle=bluebg, colframe=bluefg, fonttitle=\bfseries\color{black}, float*=ht, width=\textwidth]
\begin{Verbatim}[breaklines=true, breakanywhere=true, fontsize=\footnotesize]
</action>

## Workspace

The code and annotation of the current workflow you are working on are presented as follows:

<code>
model_15, clip_vision_15, vae_15 = ImageOnlyCheckpointLoader(ckpt_name="""svd_xt_1_1.safetensors""")
model_16, clip_16, vae_16 = CheckpointLoaderSimple(ckpt_name="""sd_xl_base_1.0.safetensors""")
conditioning_18 = CLIPTextEncode(text="""photograph beautiful scenery nature mountains alps river rapids snow sky cumulus clouds""", speak_and_recognation=True, clip=clip_16)
conditioning_19 = CLIPTextEncode(text="""text, watermark""", speak_and_recognation=True, clip=clip_16)
latent_22 = EmptyLatentImage(width=1024, height=576, batch_size=1)
model_14 = VideoLinearCFGGuidance(min_cfg=1, model=model_15)
latent_17 = KSampler(seed=307393744025667, steps=15, cfg=8, sampler_name="""uni_pc_bh2""", scheduler="""normal""", denoise=1, model=model_16, positive=conditioning_18, negative=conditioning_19, latent_image=latent_22)
image_20 = VAEDecode(samples=latent_17, vae=vae_16)
_ = PreviewImage(images=image_20)
positive_12, negative_12, latent_12 = SVD_img2vid_Conditioning(width=1024, height=576, video_frames=24, motion_bucket_id=127, fps=8, augmentation_level=0, clip_vision=clip_vision_15, init_image=image_20, vae=vae_15)
latent_3 = KSampler(seed=1103641334004632, steps=20, cfg=2.5, sampler_name="""euler""", scheduler="""karras""", denoise=1, model=model_14, positive=positive_12, negative=negative_12, latent_image=latent_12)
image_8 = VAEDecode(samples=latent_3, vae=vae_15)
filenames_23 = VHS_VideoCombine(frame_rate=6, loop_count=0, filename_prefix="""SVD_txt2vid""", format="""video/h264-mp4""", pix_fmt="""yuv420p""", crf=19, save_metadata=True, pingpong=False, save_output=True, images=image_8)

</code>

<function>
This workflow follows a text-to-video paradigm, where it generates a video from a given text prompt. It first creates an initial image based on the text description and then uses that image as the starting frame to generate a video with motion and transformation using a Stable Video Diffusion model. The output is a 3-second video at 8 frames per second, depicting beautiful scenery with mountains, rivers, and clouds.
</function>

<principle>
The workflow first loads two models: a Stable Diffusion model for generating the initial image from the text ("sd_xl_base_1.0.safetensors") and a Stable Video Diffusion model ("svd_xt_1_1.safetensors") for video generation. It uses the text description to create conditioning, generating an initial 1024x576 image. The image is decoded from latent space via a VAE. The video-specific node ("SVD_img2vid_Conditioning") then applies continuity and motion to this image, producing conditioned latent representations for video generation. The final video is created by sampling the latent space over multiple frames and combining them into an MP4 video using the specified frame rate and format.
</principle>

## Action

Based on the history and workspace, you should first think about what functions have been implemented and what modules remain to be added. Your thought should be enclosed with "<thought>" tag. For example: <thought>The basic pipeline has been implemented, but a module is needed to improve the quality.</thought>.

After that, you should update your step-by-step plan to further modify your workflow. There are 4 steps remaining, so your plan should contain at most 4 steps. Make sure that each step is feasible to be converted into a single action. Your plan should be enclosed with "<plan>" tag. For example: <plan>Step 1: I will refer to "reference_name" to add a module. Step 2: I will finish the task since the expected effects are realized.</plan>.
\end{Verbatim}
\end{tcolorbox}

\clearpage

\begin{tcolorbox}[breakable, colback=bluebg, colbacktitle=bluebg, colframe=bluefg, fonttitle=\bfseries\color{black}, float*=ht, width=\textwidth]
\begin{Verbatim}[breaklines=true, breakanywhere=true, fontsize=\footnotesize]

Finally, you should choose one of the following actions and specify the arguments (if required), so that the updated workflow can realize the first step in your plan. You should provide your action with the format of function calls in Python. Your action should be enclosed with "<action>" tag. For example: <action>combine(name="reference_name")</action>, <action>adapt(prompt="Change the factor to 0.5 and rewrite the prompt.")</action>, and <action>finish()</action>.

- `load`: Load a reference workflow into the workspace to replace the current workflow, so that you can start over. Arguments:
  - `name`: The name of the reference workflow you want to load.
- `combine`: Combine the current workflow with a reference workflow, so that necessary modules can be added. Arguments:
  - `name`: The name of the reference workflow you want to combine.
- `adapt`: Adapt some parameters in the current workflow, so that the expected effects can be realized. Arguments:
  - `prompt`: The prompt to specify the adaptation you want to make.
- `retrieve`: Retrieve a new batch of reference workflows, so that more useful references can be found. Arguments:
  - `prompt`: The prompt to describe the reference you want to retrieve.
- `finish`: Finish the task since the current workflow can realize the expected effects.

Refer to the history before making a decision. Here are some general rules you should follow:

1. You should choose the `load` action if and only if the history is empty.
2. If you choose the `load` or `combine` action, make sure the name exists in the reference. Otherwise, try to update the reference with the `retrieve` action.
3. You should not choose the `adapt` action twice in a row, because they can be simplified into a single action.
4. If you choose the `adapt` or `retrieve` action, make sure the prompt is concise and contains all the necessary information.
5. You should choose the `finish` action before the remaining steps count down to 0.

Now, provide your thought, plan and action with the required format.
\end{Verbatim}
\end{tcolorbox}

\begin{tcolorbox}[title=PlanAgent Answer, breakable, colback=redbg, colbacktitle=redbg, colframe=redfg, fonttitle=\bfseries\color{black}, float*=ht, width=\textwidth]
\begin{Verbatim}[breaklines=true, breakanywhere=true, fontsize=\footnotesize]
<thought>
The current workflow generates a 3-second video of a landscape scene, but this is not aligned with the task of generating a 2-second video of colorful fish swimming in an aquarium with bubbles rising. I need to adapt the text prompt accordingly to match the requirements. Additionally, the frame rate is currently 8 frames per second, but after generating the video, we will need to interpolate it to increase the frame rate by 3x to make the motion smoother.

The basic pipeline for generating videos from a text prompt is already functional, but the correct text needs to be provided, and the next step would be to apply video frame interpolation. I will combine this workflow with the "video_frame_interpolation" to achieve smoother motion.
</thought>

<plan>
Step 1: Adapt the text prompt to describe the required scene of "colorful fish swimming in an aquarium with bubbles rising".  
Step 2: Combine the current workflow with the "video_frame_interpolation" reference workflow to add video interpolation functionality.  
Step 3: Adapt the interpolation multiplier to 3x for smoother motion.  
Step 4: Finish the task after confirming that the workflow meets the requirements.
</plan>

<action>
adapt(prompt="Change the text prompt to 'colorful fish swimming in an aquarium with bubbles rising to the surface' and reduce the video duration to 2 seconds.")
\end{Verbatim}
\end{tcolorbox}

\clearpage

\begin{tcolorbox}[breakable, colback=redbg, colbacktitle=redbg, colframe=redfg, fonttitle=\bfseries\color{black}, float*=ht, width=\textwidth]
\begin{Verbatim}[breaklines=true, breakanywhere=true, fontsize=\footnotesize]
</action>
\end{Verbatim}
\end{tcolorbox}

\begin{tcolorbox}[title=AdaptAgent Prompt, breakable, colback=bluebg, colbacktitle=bluebg, colframe=bluefg, fonttitle=\bfseries\color{black}, float*=ht, width=\textwidth]
\begin{Verbatim}[breaklines=true, breakanywhere=true, fontsize=\footnotesize]
## Task

ComfyUI uses workflows to create and execute Stable Diffusion pipelines so that users can design their own workflows to generate highly customized artworks. ComfyUI workflows can be formulated into the equivalent Python code, where each statement represents the execution of a single node. You are an expert in ComfyUI, helping users to design workflows according to their requirements.

Now you are required to create a ComfyUI workflow to finish the following task:

First generate a 2-second video of colorful fish swimming in an aquarium with bubbles rising to the surface. Then interpolate the video to increase the frame rate by 3x. The result should be a smoother video of the swimming fish.

The core requirements and the expected paradigm are analyzed as follows:

The task requires generating a short 2-second video of fish swimming in an aquarium with bubbles rising using a text-to-video approach, as the primary input is textual. The video needs to depict dynamic, colorful fish movements with the additional visual element of rising bubbles. After generating this video, the frame rate needs to be increased by 3x using a video interpolation process to achieve smoother motion. Thus, the workflow core involves two stages, first, synthesizing the video from a text prompt using a text-to-video node or series of nodes, followed by applying video interpolation to enhance frame smoothness. There are no specific quality-oriented steps like upscaling or refinement beyond the interpolation requirement.

## Workspace

The code and annotation of the current workflow you are working on are presented as follows:

<code>
model_15, clip_vision_15, vae_15 = ImageOnlyCheckpointLoader(ckpt_name="""svd_xt_1_1.safetensors""")
model_16, clip_16, vae_16 = CheckpointLoaderSimple(ckpt_name="""sd_xl_base_1.0.safetensors""")
conditioning_18 = CLIPTextEncode(text="""photograph beautiful scenery nature mountains alps river rapids snow sky cumulus clouds""", speak_and_recognation=True, clip=clip_16)
conditioning_19 = CLIPTextEncode(text="""text, watermark""", speak_and_recognation=True, clip=clip_16)
latent_22 = EmptyLatentImage(width=1024, height=576, batch_size=1)
model_14 = VideoLinearCFGGuidance(min_cfg=1, model=model_15)
latent_17 = KSampler(seed=307393744025667, steps=15, cfg=8, sampler_name="""uni_pc_bh2""", scheduler="""normal""", denoise=1, model=model_16, positive=conditioning_18, negative=conditioning_19, latent_image=latent_22)
image_20 = VAEDecode(samples=latent_17, vae=vae_16)
_ = PreviewImage(images=image_20)
positive_12, negative_12, latent_12 = SVD_img2vid_Conditioning(width=1024, height=576, video_frames=24, motion_bucket_id=127, fps=8, augmentation_level=0, clip_vision=clip_vision_15, init_image=image_20, vae=vae_15)
latent_3 = KSampler(seed=1103641334004632, steps=20, cfg=2.5, sampler_name="""euler""", scheduler="""karras""", denoise=1, model=model_14, positive=positive_12, negative=negative_12, latent_image=latent_12)
image_8 = VAEDecode(samples=latent_3, vae=vae_15)
filenames_23 = VHS_VideoCombine(frame_rate=6, loop_count=0, filename_prefix="""SVD_txt2vid""", format="""video/h264-mp4""", pix_fmt="""yuv420p""", crf=19, save_metadata=True, pingpong=False, save_output=True, images=image_8)

</code>

<function>
\end{Verbatim}
\end{tcolorbox}

\clearpage

\begin{tcolorbox}[breakable, colback=bluebg, colbacktitle=bluebg, colframe=bluefg, fonttitle=\bfseries\color{black}, float*=ht, width=\textwidth]
\begin{Verbatim}[breaklines=true, breakanywhere=true, fontsize=\footnotesize]
This workflow follows a text-to-video paradigm, where it generates a video from a given text prompt. It first creates an initial image based on the text description and then uses that image as the starting frame to generate a video with motion and transformation using a Stable Video Diffusion model. The output is a 3-second video at 8 frames per second, depicting beautiful scenery with mountains, rivers, and clouds.
</function>

<principle>
The workflow first loads two models: a Stable Diffusion model for generating the initial image from the text ("sd_xl_base_1.0.safetensors") and a Stable Video Diffusion model ("svd_xt_1_1.safetensors") for video generation. It uses the text description to create conditioning, generating an initial 1024x576 image. The image is decoded from latent space via a VAE. The video-specific node ("SVD_img2vid_Conditioning") then applies continuity and motion to this image, producing conditioned latent representations for video generation. The final video is created by sampling the latent space over multiple frames and combining them into an MP4 video using the specified frame rate and format.
</principle>

## Adaptation

Based on the current working progress, your schedule is presented as follows:

Step 1: Adapt the text prompt to describe the required scene of "colorful fish swimming in an aquarium with bubbles rising".  
Step 2: Combine the current workflow with the "video_frame_interpolation" reference workflow to add video interpolation functionality.  
Step 3: Adapt the interpolation multiplier to 3x for smoother motion.  
Step 4: Finish the task after confirming that the workflow meets the requirements.

You are working on the first step of your schedule. In other words, you should modify the parameters in the current workflow according to your schedule. The adaptation you want to make is specified as follows:

Change the text prompt to 'colorful fish swimming in an aquarium with bubbles rising to the surface' and reduce the video duration to 2 seconds.

First, you should provide your Python code to formulate the updated workflow. Each line of code should correspond to a single node, so you should avoid nested calls in a single statement. You should also avoid reusing the same variable name, even if the variable is temporary. Your code should be enclosed with "<code>" tag. For example: <code>output = node(input)</code>.

After that, you should provide an annotation as in the reference, including the function and principle of the updated workflow. The function should be enclosed with "<function>" tag. For example: <function>This workflow generates a high-resolution image of a running horse.</function>. The principle should be enclosed with "<principle>" tag. For example: <principle>The workflow first generates a low-resolution image using the text-to-image pipeline and then applies an upscaling module to improve the resolution.</principle>.

Now, provide your code and annotation with the required format.
\end{Verbatim}
\end{tcolorbox}

\begin{tcolorbox}[title=AdaptAgent Answer, breakable, colback=redbg, colbacktitle=redbg, colframe=redfg, fonttitle=\bfseries\color{black}, float*=ht, width=\textwidth]
\begin{Verbatim}[breaklines=true, breakanywhere=true, fontsize=\footnotesize]
<code>
model_15, clip_vision_15, vae_15 = ImageOnlyCheckpointLoader(ckpt_name="""svd_xt_1_1.safetensors""")  
model_16, clip_16, vae_16 = CheckpointLoaderSimple(ckpt_name="""sd_xl_base_1.0.safetensors""")  
conditioning_18 = CLIPTextEncode(text="""colorful fish swimming in an aquarium with bubbles rising to the surface""", speak_and_recognation=True, clip=clip_16)  
conditioning_19 = CLIPTextEncode(text="""text, watermark""", speak_and_recognation=True, clip=clip_16)  
latent_22 = EmptyLatentImage(width=1024, height=576, batch_size=1)  
model_14 = VideoLinearCFGGuidance(min_cfg=1, model=model_15)  
\end{Verbatim}
\end{tcolorbox}

\clearpage

\begin{tcolorbox}[breakable, colback=redbg, colbacktitle=redbg, colframe=redfg, fonttitle=\bfseries\color{black}, float*=ht, width=\textwidth]
\begin{Verbatim}[breaklines=true, breakanywhere=true, fontsize=\footnotesize]
latent_17 = KSampler(seed=307393744025667, steps=15, cfg=8, sampler_name="""uni_pc_bh2""", scheduler="""normal""", denoise=1, model=model_16, positive=conditioning_18, negative=conditioning_19, latent_image=latent_22)  
image_20 = VAEDecode(samples=latent_17, vae=vae_16)  
_ = PreviewImage(images=image_20)  
positive_12, negative_12, latent_12 = SVD_img2vid_Conditioning(width=1024, height=576, video_frames=16, motion_bucket_id=127, fps=8, augmentation_level=0, clip_vision=clip_vision_15, init_image=image_20, vae=vae_15)  
latent_3 = KSampler(seed=1103641334004632, steps=20, cfg=2.5, sampler_name="""euler""", scheduler="""karras""", denoise=1, model=model_14, positive=positive_12, negative=negative_12, latent_image=latent_12)  
image_8 = VAEDecode(samples=latent_3, vae=vae_15)  
filenames_23 = VHS_VideoCombine(frame_rate=8, loop_count=0, filename_prefix="""SVD_txt2vid""", format="""video/h264-mp4""", pix_fmt="""yuv420p""", crf=19, save_metadata=True, pingpong=False, save_output=True, images=image_8)  
</code>

<function>
This workflow generates a 2-second video of colorful fish swimming in an aquarium with bubbles rising to the surface from a given text description. The video is first synthesized using a text-to-video diffusion process.
</function>

<principle>
The workflow first loads two models: a Stable Diffusion model ("sd_xl_base_1.0.safetensors") for generating the initial image from the text description, and then a Stable Video Diffusion model ("svd_xt_1_1.safetensors") to translate that image into a dynamic video over multiple frames. Following the text input describing colorful fish and bubbles, the image is generated and expanded across 16 frames, which corresponds to a duration of 2 seconds at 8 FPS (frames per second). The final MP4 video is constructed and saved along with metadata.
</principle>
\end{Verbatim}
\end{tcolorbox}

\begin{tcolorbox}[title=PlanAgent Prompt, breakable, colback=bluebg, colbacktitle=bluebg, colframe=bluefg, fonttitle=\bfseries\color{black}, float*=ht, width=\textwidth]
\begin{Verbatim}[breaklines=true, breakanywhere=true, fontsize=\footnotesize]
## Task

ComfyUI uses workflows to create and execute Stable Diffusion pipelines so that users can design their own workflows to generate highly customized artworks. ComfyUI workflows can be formulated into the equivalent Python code, where each statement represents the execution of a single node. You are an expert in ComfyUI, helping users to design workflows according to their requirements.

Now you are required to create a ComfyUI workflow to finish the following task:

First generate a 2-second video of colorful fish swimming in an aquarium with bubbles rising to the surface. Then interpolate the video to increase the frame rate by 3x. The result should be a smoother video of the swimming fish.

The core requirements and the expected paradigm are analyzed as follows:

The task requires generating a short 2-second video of fish swimming in an aquarium with bubbles rising using a text-to-video approach, as the primary input is textual. The video needs to depict dynamic, colorful fish movements with the additional visual element of rising bubbles. After generating this video, the frame rate needs to be increased by 3x using a video interpolation process to achieve smoother motion. Thus, the workflow core involves two stages, first, synthesizing the video from a text prompt using a text-to-video node or series of nodes, followed by applying video interpolation to enhance frame smoothness. There are no specific quality-oriented steps like upscaling or refinement beyond the interpolation requirement.

Improving the quality of the generation result with additional steps, such as upscaling and interpolation, is not recommended, unless specified in the requirements.

\end{Verbatim}
\end{tcolorbox}

\clearpage

\begin{tcolorbox}[breakable, colback=bluebg, colbacktitle=bluebg, colframe=bluefg, fonttitle=\bfseries\color{black}, float*=ht, width=\textwidth]
\begin{Verbatim}[breaklines=true, breakanywhere=true, fontsize=\footnotesize]
## Reference

According to the requirements, we have retrieved some relevant workflows which may be helpful:

- Example: text_to_video

<function>
This workflow follows a text-to-video paradigm, where it generates a video from a given text prompt. It first creates an initial image based on the text description and then uses that image as the starting frame to generate a video with motion and transformation using a Stable Video Diffusion model. The output is a 3-second video at 8 frames per second, depicting beautiful scenery with mountains, rivers, and clouds.
</function>

<principle>
The workflow first loads two models: a Stable Diffusion model for generating the initial image from the text ("sd_xl_base_1.0.safetensors") and a Stable Video Diffusion model ("svd_xt_1_1.safetensors") for video generation. It uses the text description to create conditioning, generating an initial 1024x576 image. The image is decoded from latent space via a VAE. The video-specific node ("SVD_img2vid_Conditioning") then applies continuity and motion to this image, producing conditioned latent representations for video generation. The final video is created by sampling the latent space over multiple frames and combining them into an MP4 video using the specified frame rate and format.
</principle>

- Example: video_frame_interpolation

<function>
This workflow performs video frame interpolation using the RIFE VFI model. It takes an input video such as "play_guitar.gif", increases the frame rate by generating intermediate frames (interpolating) with a multiplier (in this case, 3x), and produces a smoother video with a higher frame rate (from 8 to 24 frames per second). The final output is saved as a new video or animated GIF.
</function>

<principle>
The workflow first loads the input video using "VHS_LoadVideo", which extracts the individual frames. The "RIFE VFI" node is then used to interpolate the frames by generating additional frames between the existing ones. In this scenario, the multiplier is set to 3x, effectively tripling the frame count and enabling a smoother video playback at 24 frames per second. Finally, the interpolated frames are combined into a video or GIF format using "VHS_VideoCombine".
</principle>

- Example: image_to_video

<function>
This workflow follows an image-to-video paradigm. It requires an input image (in this case, "play_guitar.jpg") and generates a 4-second video at 6 frames per second (24 video frames in total) based on that image. The workflow outputs the generated video.
</function>

<principle>
The workflow uses the "svd_xt_1_1.safetensors" Stable Video Diffusion model to generate a video from the input image "play_guitar.jpg". The "SVD_img2vid_Conditioning" node creates the necessary conditioning for video generation, including the number of frames, resolution, and motion characteristics. A KSamplerAdvanced node adds noise and performs generative sampling over multiple steps to create diverse video frames. These frames are then decoded back into images via a VAE, and finally, the "VHS_VideoCombine" node compiles these images into a 4-second video at 8 frames per second.
</principle>

- Example: text_to_image

<function>
\end{Verbatim}
\end{tcolorbox}

\clearpage

\begin{tcolorbox}[breakable, colback=bluebg, colbacktitle=bluebg, colframe=bluefg, fonttitle=\bfseries\color{black}, float*=ht, width=\textwidth]
\begin{Verbatim}[breaklines=true, breakanywhere=true, fontsize=\footnotesize]
This workflow implements a basic text-to-image generation pipeline using Stable Diffusion. It requires both positive (desired traits) and negative (undesired traits) text prompts to generate an image. In this specific case, the workflow will output a high-resolution photo of a cat wearing a spacesuit inside a spaceship, avoiding blurry or illustration-like effects.
</function>

<principle>
The workflow begins by loading the "dreamshaper_8.safetensors" Stable Diffusion model. It generates a blank latent space as the starting point for the image generation. The positive prompt ("a photo of a cat wearing a spacesuit inside a spaceship") and negative prompt ("blurry, illustration") are encoded into conditioning by the CLIPTextEncode node. The KSampler node then uses these conditionings to guide the generative process over 20 sampling steps, applying denoising to the latent space. The resulting latent code is subsequently decoded into an image using the VAE and saved to disk.
</principle>

- Example: scribble_image_repaint

<function>
This workflow follows a sketch-to-image paradigm, where it takes a scribble image ("simple_graffiti.png") along with a text prompt ("a bird, open wings") and generates a detailed, high-quality image based on both the scribble and the prompt.
</function>

<principle>
The workflow first loads and inverts the input scribble image. The inverted image is then used by a "control_v11p_sd15_scribble_fp16" ControlNet model to extract and applied to the scribble, controlling and guiding the image generation process. A pre-trained model ("dreamshaper_8.safetensors") processes the positive conditioning generated by the combination of the text prompt and the ControlNet's output, along with negative conditioning from undesired traits such as "horror" and "lowres." The latent space image is generated using a KSampler and further decoded by the VAE to form the final image. The model synthesizes a highly detailed image while adhering closely to both the shape of the scribble and the desired features implied by the text prompt.
</principle>

## History

Here is a recent history of your thought, plan and action in the previous steps. The most recent record is at the bottom.

- Step: 1

<thought>
Since the history is empty, I need to start by loading a relevant reference workflow that can be used as a base for generating the 2-second video of colorful fish in an aquarium. The "text_to_video" reference seems appropriate to use as it involves generating video from text prompts, which aligns with the first part of the task.
</thought>

<plan>
Step 1: I will load the "text_to_video" reference workflow to use it as the base.  
Step 2: I will adapt the workflow to modify the prompt to match the requirements ("colorful fish swimming in an aquarium with bubbles rising").  
Step 3: I will combine it with the "video_frame_interpolation" reference to enable interpolation for smoother video.  
Step 4: I will adapt the interpolation multiplier to 3x to achieve the expected frame rate increase.  
Step 5: I will finish the task by confirming the generated output meets the required specifications.
</plan>

<action>
load(name="text_to_video")
\end{Verbatim}
\end{tcolorbox}

\clearpage

\begin{tcolorbox}[breakable, colback=bluebg, colbacktitle=bluebg, colframe=bluefg, fonttitle=\bfseries\color{black}, float*=ht, width=\textwidth]
\begin{Verbatim}[breaklines=true, breakanywhere=true, fontsize=\footnotesize]
</action>

- Step: 2

<thought>
The current workflow generates a 3-second video of a landscape scene, but this is not aligned with the task of generating a 2-second video of colorful fish swimming in an aquarium with bubbles rising. I need to adapt the text prompt accordingly to match the requirements. Additionally, the frame rate is currently 8 frames per second, but after generating the video, we will need to interpolate it to increase the frame rate by 3x to make the motion smoother.

The basic pipeline for generating videos from a text prompt is already functional, but the correct text needs to be provided, and the next step would be to apply video frame interpolation. I will combine this workflow with the "video_frame_interpolation" to achieve smoother motion.
</thought>

<plan>
Step 1: Adapt the text prompt to describe the required scene of "colorful fish swimming in an aquarium with bubbles rising".  
Step 2: Combine the current workflow with the "video_frame_interpolation" reference workflow to add video interpolation functionality.  
Step 3: Adapt the interpolation multiplier to 3x for smoother motion.  
Step 4: Finish the task after confirming that the workflow meets the requirements.
</plan>

<action>
adapt(prompt="Change the text prompt to 'colorful fish swimming in an aquarium with bubbles rising to the surface' and reduce the video duration to 2 seconds.")
</action>

## Workspace

The code and annotation of the current workflow you are working on are presented as follows:

<code>
model_15, clip_vision_15, vae_15 = ImageOnlyCheckpointLoader(ckpt_name="""svd_xt_1_1.safetensors""")  
model_16, clip_16, vae_16 = CheckpointLoaderSimple(ckpt_name="""sd_xl_base_1.0.safetensors""")  
conditioning_18 = CLIPTextEncode(text="""colorful fish swimming in an aquarium with bubbles rising to the surface""", speak_and_recognation=True, clip=clip_16)  
conditioning_19 = CLIPTextEncode(text="""text, watermark""", speak_and_recognation=True, clip=clip_16)  
latent_22 = EmptyLatentImage(width=1024, height=576, batch_size=1)  
model_14 = VideoLinearCFGGuidance(min_cfg=1, model=model_15)  
latent_17 = KSampler(seed=307393744025667, steps=15, cfg=8, sampler_name="""uni_pc_bh2""", scheduler="""normal""", denoise=1, model=model_16, positive=conditioning_18, negative=conditioning_19, latent_image=latent_22)  
image_20 = VAEDecode(samples=latent_17, vae=vae_16)  
_ = PreviewImage(images=image_20)  
positive_12, negative_12, latent_12 = SVD_img2vid_Conditioning(width=1024, height=576, video_frames=16, motion_bucket_id=127, fps=8, augmentation_level=0, clip_vision=clip_vision_15, init_image=image_20, vae=vae_15)  
latent_3 = KSampler(seed=1103641334004632, steps=20, cfg=2.5, sampler_name="""euler""", scheduler="""karras""", denoise=1, model=model_14, positive=positive_12, negative=negative_12, latent_image=latent_12)  
image_8 = VAEDecode(samples=latent_3, vae=vae_15)  
filenames_23 = VHS_VideoCombine(frame_rate=8, loop_count=0, filename_prefix="""SVD_txt2vid""", format="""video/h264-mp4""", pix_fmt="""yuv420p""", crf=19, save_metadata=True, pingpong=False, save_output=True, images=image_8)
</code>

<function>
\end{Verbatim}
\end{tcolorbox}

\clearpage

\begin{tcolorbox}[breakable, colback=bluebg, colbacktitle=bluebg, colframe=bluefg, fonttitle=\bfseries\color{black}, float*=ht, width=\textwidth]
\begin{Verbatim}[breaklines=true, breakanywhere=true, fontsize=\footnotesize]
This workflow generates a 2-second video of colorful fish swimming in an aquarium with bubbles rising to the surface from a given text description. The video is first synthesized using a text-to-video diffusion process.
</function>

<principle>
The workflow first loads two models: a Stable Diffusion model ("sd_xl_base_1.0.safetensors") for generating the initial image from the text description, and then a Stable Video Diffusion model ("svd_xt_1_1.safetensors") to translate that image into a dynamic video over multiple frames. Following the text input describing colorful fish and bubbles, the image is generated and expanded across 16 frames, which corresponds to a duration of 2 seconds at 8 FPS (frames per second). The final MP4 video is constructed and saved along with metadata.
</principle>

## Action

Based on the history and workspace, you should first think about what functions have been implemented and what modules remain to be added. Your thought should be enclosed with "<thought>" tag. For example: <thought>The basic pipeline has been implemented, but a module is needed to improve the quality.</thought>.

After that, you should update your step-by-step plan to further modify your workflow. There are 3 steps remaining, so your plan should contain at most 3 steps. Make sure that each step is feasible to be converted into a single action. Your plan should be enclosed with "<plan>" tag. For example: <plan>Step 1: I will refer to "reference_name" to add a module. Step 2: I will finish the task since the expected effects are realized.</plan>.

Finally, you should choose one of the following actions and specify the arguments (if required), so that the updated workflow can realize the first step in your plan. You should provide your action with the format of function calls in Python. Your action should be enclosed with "<action>" tag. For example: <action>combine(name="reference_name")</action>, <action>adapt(prompt="Change the factor to 0.5 and rewrite the prompt.")</action>, and <action>finish()</action>.

- `load`: Load a reference workflow into the workspace to replace the current workflow, so that you can start over. Arguments:
  - `name`: The name of the reference workflow you want to load.
- `combine`: Combine the current workflow with a reference workflow, so that necessary modules can be added. Arguments:
  - `name`: The name of the reference workflow you want to combine.
- `adapt`: Adapt some parameters in the current workflow, so that the expected effects can be realized. Arguments:
  - `prompt`: The prompt to specify the adaptation you want to make.
- `retrieve`: Retrieve a new batch of reference workflows, so that more useful references can be found. Arguments:
  - `prompt`: The prompt to describe the reference you want to retrieve.
- `finish`: Finish the task since the current workflow can realize the expected effects.

Refer to the history before making a decision. Here are some general rules you should follow:

1. You should choose the `load` action if and only if the history is empty.
2. If you choose the `load` or `combine` action, make sure the name exists in the reference. Otherwise, try to update the reference with the `retrieve` action.
3. You should not choose the `adapt` action twice in a row, because they can be simplified into a single action.
4. If you choose the `adapt` or `retrieve` action, make sure the prompt is concise and contains all the necessary information.
5. You should choose the `finish` action before the remaining steps count down to 0.

Now, provide your thought, plan and action with the required format.
\end{Verbatim}
\end{tcolorbox}

\clearpage

\begin{tcolorbox}[title=PlanAgent Answer, breakable, colback=redbg, colbacktitle=redbg, colframe=redfg, fonttitle=\bfseries\color{black}, float*=ht, width=\textwidth]
\begin{Verbatim}[breaklines=true, breakanywhere=true, fontsize=\footnotesize]
<thought>  
The video generation part is complete, wherein a 2-second video of colorful fish swimming in an aquarium with bubbles rising is generated at 8 FPS. The next part of the task is to increase the frame rate by 3x using frame interpolation to achieve smoother motion. I need to combine the current workflow with the "video_frame_interpolation" reference to add the interpolation capability and then adjust the frame rate multiplier to 3x as per the requirements.  
</thought>

<plan>  
Step 1: Combine the current workflow with the "video_frame_interpolation" reference to add the interpolation functionality.  
Step 2: Adapt the multiplier for frame interpolation to 3x to ensure the frame rate increase.  
Step 3: Finish the task since this will meet the stated requirements.  
</plan>

<action>combine(name="video_frame_interpolation")</action>  
\end{Verbatim}
\end{tcolorbox}

\begin{tcolorbox}[title=CombineAgent Prompt, breakable, colback=bluebg, colbacktitle=bluebg, colframe=bluefg, fonttitle=\bfseries\color{black}, float*=ht, width=\textwidth]
\begin{Verbatim}[breaklines=true, breakanywhere=true, fontsize=\footnotesize]
## Task

ComfyUI uses workflows to create and execute Stable Diffusion pipelines so that users can design their own workflows to generate highly customized artworks. ComfyUI workflows can be formulated into the equivalent Python code, where each statement represents the execution of a single node. You are an expert in ComfyUI, helping users to design workflows according to their requirements.

Now you are required to create a ComfyUI workflow to finish the following task:

First generate a 2-second video of colorful fish swimming in an aquarium with bubbles rising to the surface. Then interpolate the video to increase the frame rate by 3x. The result should be a smoother video of the swimming fish.

The core requirements and the expected paradigm are analyzed as follows:

The task requires generating a short 2-second video of fish swimming in an aquarium with bubbles rising using a text-to-video approach, as the primary input is textual. The video needs to depict dynamic, colorful fish movements with the additional visual element of rising bubbles. After generating this video, the frame rate needs to be increased by 3x using a video interpolation process to achieve smoother motion. Thus, the workflow core involves two stages, first, synthesizing the video from a text prompt using a text-to-video node or series of nodes, followed by applying video interpolation to enhance frame smoothness. There are no specific quality-oriented steps like upscaling or refinement beyond the interpolation requirement.

## Reference

The code and annotation of the current workflow you are referring to are presented as follows:

<code>
image_7, frame_count_7, audio_7, video_info_7, latent_7 = VHS_LoadVideo(custom_height=512, custom_width=512, force_rate=0, force_size="""Disabled""", frame_load_cap=0, select_every_nth=1, skip_first_frames=0, video="""play_guitar.gif""")
image_10 = RIFE_VFI(ckpt_name="""rife47.pth""", clear_cache_after_n_frames=10, ensemble=True, fast_mode=True, frames=image_7, multiplier=3, scale_factor=1)
filenames_3 = VHS_VideoCombine(filename_prefix="""AnimateDiff""", format="""image/gif""", frame_rate=24, images=image_10, loop_count=0, pingpong=False, save_output=True)
</code>

<function>
\end{Verbatim}
\end{tcolorbox}

\clearpage

\begin{tcolorbox}[breakable, colback=bluebg, colbacktitle=bluebg, colframe=bluefg, fonttitle=\bfseries\color{black}, float*=ht, width=\textwidth]
\begin{Verbatim}[breaklines=true, breakanywhere=true, fontsize=\footnotesize]
This workflow performs video frame interpolation using the RIFE VFI model. It takes an input video such as "play_guitar.gif", increases the frame rate by generating intermediate frames (interpolating) with a multiplier (in this case, 3x), and produces a smoother video with a higher frame rate (from 8 to 24 frames per second). The final output is saved as a new video or animated GIF.
</function>

<principle>
The workflow first loads the input video using "VHS_LoadVideo", which extracts the individual frames. The "RIFE VFI" node is then used to interpolate the frames by generating additional frames between the existing ones. In this scenario, the multiplier is set to 3x, effectively tripling the frame count and enabling a smoother video playback at 24 frames per second. Finally, the interpolated frames are combined into a video or GIF format using "VHS_VideoCombine".
</principle>

## Workspace

The code and annotation of the current workflow you are working on are presented as follows:

<code>
model_15, clip_vision_15, vae_15 = ImageOnlyCheckpointLoader(ckpt_name="""svd_xt_1_1.safetensors""")  
model_16, clip_16, vae_16 = CheckpointLoaderSimple(ckpt_name="""sd_xl_base_1.0.safetensors""")  
conditioning_18 = CLIPTextEncode(text="""colorful fish swimming in an aquarium with bubbles rising to the surface""", speak_and_recognation=True, clip=clip_16)  
conditioning_19 = CLIPTextEncode(text="""text, watermark""", speak_and_recognation=True, clip=clip_16)  
latent_22 = EmptyLatentImage(width=1024, height=576, batch_size=1)  
model_14 = VideoLinearCFGGuidance(min_cfg=1, model=model_15)  
latent_17 = KSampler(seed=307393744025667, steps=15, cfg=8, sampler_name="""uni_pc_bh2""", scheduler="""normal""", denoise=1, model=model_16, positive=conditioning_18, negative=conditioning_19, latent_image=latent_22)  
image_20 = VAEDecode(samples=latent_17, vae=vae_16)  
_ = PreviewImage(images=image_20)  
positive_12, negative_12, latent_12 = SVD_img2vid_Conditioning(width=1024, height=576, video_frames=16, motion_bucket_id=127, fps=8, augmentation_level=0, clip_vision=clip_vision_15, init_image=image_20, vae=vae_15)  
latent_3 = KSampler(seed=1103641334004632, steps=20, cfg=2.5, sampler_name="""euler""", scheduler="""karras""", denoise=1, model=model_14, positive=positive_12, negative=negative_12, latent_image=latent_12)  
image_8 = VAEDecode(samples=latent_3, vae=vae_15)  
filenames_23 = VHS_VideoCombine(frame_rate=8, loop_count=0, filename_prefix="""SVD_txt2vid""", format="""video/h264-mp4""", pix_fmt="""yuv420p""", crf=19, save_metadata=True, pingpong=False, save_output=True, images=image_8)
</code>

<function>
This workflow generates a 2-second video of colorful fish swimming in an aquarium with bubbles rising to the surface from a given text description. The video is first synthesized using a text-to-video diffusion process.
</function>

<principle>
The workflow first loads two models: a Stable Diffusion model ("sd_xl_base_1.0.safetensors") for generating the initial image from the text description, and then a Stable Video Diffusion model ("svd_xt_1_1.safetensors") to translate that image into a dynamic video over multiple frames. Following the text input describing colorful fish and bubbles, the image is generated and expanded across 16 frames, which corresponds to a duration of 2 seconds at 8 FPS (frames per second). The final MP4 video is constructed and saved along with metadata.
</principle>

## Combination

Based on the current working progress, your schedule is presented as follows:
\end{Verbatim}
\end{tcolorbox}

\clearpage

\begin{tcolorbox}[breakable, colback=bluebg, colbacktitle=bluebg, colframe=bluefg, fonttitle=\bfseries\color{black}, float*=ht, width=\textwidth]
\begin{Verbatim}[breaklines=true, breakanywhere=true, fontsize=\footnotesize]

Step 1: Combine the current workflow with the "video_frame_interpolation" reference to add the interpolation functionality.  
Step 2: Adapt the multiplier for frame interpolation to 3x to ensure the frame rate increase.  
Step 3: Finish the task since this will meet the stated requirements.

You are working on the first step of your schedule. In other words, you should combine the reference workflow with the current workflow according to your schedule.

First, you should provide your Python code to formulate the updated workflow. Each line of code should correspond to a single node, so you should avoid nested calls in a single statement. You should also avoid reusing the same variable name, even if the variable is temporary. Your code should be enclosed with "<code>" tag. For example: <code>output = node(input)</code>.

After that, you should provide an annotation as in the reference, including the function and principle of the updated workflow. The function should be enclosed with "<function>" tag. For example: <function>This workflow generates a high-resolution image of a running horse.</function>. The principle should be enclosed with "<principle>" tag. For example: <principle>The workflow first generates a low-resolution image using the text-to-image pipeline and then applies an upscaling module to improve the resolution.</principle>.

Now, provide your code and annotation with the required format.
\end{Verbatim}
\end{tcolorbox}

\begin{tcolorbox}[title=CombineAgent Answer, breakable, colback=redbg, colbacktitle=redbg, colframe=redfg, fonttitle=\bfseries\color{black}, float*=ht, width=\textwidth]
\begin{Verbatim}[breaklines=true, breakanywhere=true, fontsize=\footnotesize]
<code>
model_15, clip_vision_15, vae_15 = ImageOnlyCheckpointLoader(ckpt_name="""svd_xt_1_1.safetensors""")  
model_16, clip_16, vae_16 = CheckpointLoaderSimple(ckpt_name="""sd_xl_base_1.0.safetensors""")  
conditioning_18 = CLIPTextEncode(text="""colorful fish swimming in an aquarium with bubbles rising to the surface""", speak_and_recognation=True, clip=clip_16)  
conditioning_19 = CLIPTextEncode(text="""text, watermark""", speak_and_recognation=True, clip=clip_16)  
latent_22 = EmptyLatentImage(width=1024, height=576, batch_size=1)  
model_14 = VideoLinearCFGGuidance(min_cfg=1, model=model_15)  
latent_17 = KSampler(seed=307393744025667, steps=15, cfg=8, sampler_name="""uni_pc_bh2""", scheduler="""normal""", denoise=1, model=model_16, positive=conditioning_18, negative=conditioning_19, latent_image=latent_22)  
image_20 = VAEDecode(samples=latent_17, vae=vae_16)  
_ = PreviewImage(images=image_20)  
positive_12, negative_12, latent_12 = SVD_img2vid_Conditioning(width=1024, height=576, video_frames=16, motion_bucket_id=127, fps=8, augmentation_level=0, clip_vision=clip_vision_15, init_image=image_20, vae=vae_15)  
latent_3 = KSampler(seed=1103641334004632, steps=20, cfg=2.5, sampler_name="""euler""", scheduler="""karras""", denoise=1, model=model_14, positive=positive_12, negative=negative_12, latent_image=latent_12)  
image_8 = VAEDecode(samples=latent_3, vae=vae_15)  
filenames_23 = VHS_VideoCombine(frame_rate=8, loop_count=0, filename_prefix="""SVD_txt2vid""", format="""video/h264-mp4""", pix_fmt="""yuv420p""", crf=19, save_metadata=True, pingpong=False, save_output=True, images=image_8)

image_10 = RIFE_VFI(ckpt_name="""rife47.pth""", clear_cache_after_n_frames=10, ensemble=True, fast_mode=True, frames=image_8, multiplier=3, scale_factor=1)  
filenames_24 = VHS_VideoCombine(frame_rate=24, loop_count=0, filename_prefix="""SVD_txt2vid_Interpolated""", format="""video/h264-mp4""", pix_fmt="""yuv420p""", crf=19, save_metadata=True, pingpong=False, save_output=True, images=image_10)
</code>

<function>
\end{Verbatim}
\end{tcolorbox}

\clearpage

\begin{tcolorbox}[breakable, colback=redbg, colbacktitle=redbg, colframe=redfg, fonttitle=\bfseries\color{black}, float*=ht, width=\textwidth]
\begin{Verbatim}[breaklines=true, breakanywhere=true, fontsize=\footnotesize]
This workflow first generates a 2-second video of colorful fish swimming in an aquarium with bubbles rising to the surface from a text description using a text-to-video approach. It then performs video interpolation to increase the frame rate by 3x, resulting in a smoother video.
</function>

<principle>
The workflow begins by loading the necessary models, including the Stable Diffusion model and the Stable Video Diffusion model. These are used to encode the input text to create a video showcasing colorful fish and bubbles. Initially, the video is generated at 8 frames per second (FPS) over 16 frames (~2 seconds). After the video is synthesized, it is passed through a frame interpolation process using the RIFE VFI model, which increases the frame rate by generating additional frames, multiplying the frame rate by 3x (from 8 FPS to 24 FPS). Finally, the interpolated frames are combined into a smoother MP4 video output.
</principle>
\end{Verbatim}
\end{tcolorbox}

\begin{tcolorbox}[title=PlanAgent Prompt, breakable, colback=bluebg, colbacktitle=bluebg, colframe=bluefg, fonttitle=\bfseries\color{black}, float*=ht, width=\textwidth]
\begin{Verbatim}[breaklines=true, breakanywhere=true, fontsize=\footnotesize]
## Task

ComfyUI uses workflows to create and execute Stable Diffusion pipelines so that users can design their own workflows to generate highly customized artworks. ComfyUI workflows can be formulated into the equivalent Python code, where each statement represents the execution of a single node. You are an expert in ComfyUI, helping users to design workflows according to their requirements.

Now you are required to create a ComfyUI workflow to finish the following task:

First generate a 2-second video of colorful fish swimming in an aquarium with bubbles rising to the surface. Then interpolate the video to increase the frame rate by 3x. The result should be a smoother video of the swimming fish.

The core requirements and the expected paradigm are analyzed as follows:

The task requires generating a short 2-second video of fish swimming in an aquarium with bubbles rising using a text-to-video approach, as the primary input is textual. The video needs to depict dynamic, colorful fish movements with the additional visual element of rising bubbles. After generating this video, the frame rate needs to be increased by 3x using a video interpolation process to achieve smoother motion. Thus, the workflow core involves two stages, first, synthesizing the video from a text prompt using a text-to-video node or series of nodes, followed by applying video interpolation to enhance frame smoothness. There are no specific quality-oriented steps like upscaling or refinement beyond the interpolation requirement.

Improving the quality of the generation result with additional steps, such as upscaling and interpolation, is not recommended, unless specified in the requirements.

## Reference

According to the requirements, we have retrieved some relevant workflows which may be helpful:

- Example: text_to_video

<function>
This workflow follows a text-to-video paradigm, where it generates a video from a given text prompt. It first creates an initial image based on the text description and then uses that image as the starting frame to generate a video with motion and transformation using a Stable Video Diffusion model. The output is a 3-second video at 8 frames per second, depicting beautiful scenery with mountains, rivers, and clouds.
</function>

<principle>
\end{Verbatim}
\end{tcolorbox}

\clearpage

\begin{tcolorbox}[breakable, colback=bluebg, colbacktitle=bluebg, colframe=bluefg, fonttitle=\bfseries\color{black}, float*=ht, width=\textwidth]
\begin{Verbatim}[breaklines=true, breakanywhere=true, fontsize=\footnotesize]
The workflow first loads two models: a Stable Diffusion model for generating the initial image from the text ("sd_xl_base_1.0.safetensors") and a Stable Video Diffusion model ("svd_xt_1_1.safetensors") for video generation. It uses the text description to create conditioning, generating an initial 1024x576 image. The image is decoded from latent space via a VAE. The video-specific node ("SVD_img2vid_Conditioning") then applies continuity and motion to this image, producing conditioned latent representations for video generation. The final video is created by sampling the latent space over multiple frames and combining them into an MP4 video using the specified frame rate and format.
</principle>

- Example: video_frame_interpolation

<function>
This workflow performs video frame interpolation using the RIFE VFI model. It takes an input video such as "play_guitar.gif", increases the frame rate by generating intermediate frames (interpolating) with a multiplier (in this case, 3x), and produces a smoother video with a higher frame rate (from 8 to 24 frames per second). The final output is saved as a new video or animated GIF.
</function>

<principle>
The workflow first loads the input video using "VHS_LoadVideo", which extracts the individual frames. The "RIFE VFI" node is then used to interpolate the frames by generating additional frames between the existing ones. In this scenario, the multiplier is set to 3x, effectively tripling the frame count and enabling a smoother video playback at 24 frames per second. Finally, the interpolated frames are combined into a video or GIF format using "VHS_VideoCombine".
</principle>

- Example: image_to_video

<function>
This workflow follows an image-to-video paradigm. It requires an input image (in this case, "play_guitar.jpg") and generates a 4-second video at 6 frames per second (24 video frames in total) based on that image. The workflow outputs the generated video.
</function>

<principle>
The workflow uses the "svd_xt_1_1.safetensors" Stable Video Diffusion model to generate a video from the input image "play_guitar.jpg". The "SVD_img2vid_Conditioning" node creates the necessary conditioning for video generation, including the number of frames, resolution, and motion characteristics. A KSamplerAdvanced node adds noise and performs generative sampling over multiple steps to create diverse video frames. These frames are then decoded back into images via a VAE, and finally, the "VHS_VideoCombine" node compiles these images into a 4-second video at 8 frames per second.
</principle>

- Example: text_to_image

<function>
This workflow implements a basic text-to-image generation pipeline using Stable Diffusion. It requires both positive (desired traits) and negative (undesired traits) text prompts to generate an image. In this specific case, the workflow will output a high-resolution photo of a cat wearing a spacesuit inside a spaceship, avoiding blurry or illustration-like effects.
</function>

<principle>
The workflow begins by loading the "dreamshaper_8.safetensors" Stable Diffusion model. It generates a blank latent space as the starting point for the image generation. The positive prompt ("a photo of a cat wearing a spacesuit inside a spaceship") and negative prompt ("blurry, illustration") are encoded into conditioning by the CLIPTextEncode node. The KSampler node then uses these conditionings to guide the generative process over 20 sampling steps, applying denoising to the latent space. The resulting latent code is subsequently decoded into an image using the VAE and saved to disk.
\end{Verbatim}
\end{tcolorbox}

\clearpage

\begin{tcolorbox}[breakable, colback=bluebg, colbacktitle=bluebg, colframe=bluefg, fonttitle=\bfseries\color{black}, float*=ht, width=\textwidth]
\begin{Verbatim}[breaklines=true, breakanywhere=true, fontsize=\footnotesize]
</principle>

- Example: scribble_image_repaint

<function>
This workflow follows a sketch-to-image paradigm, where it takes a scribble image ("simple_graffiti.png") along with a text prompt ("a bird, open wings") and generates a detailed, high-quality image based on both the scribble and the prompt.
</function>

<principle>
The workflow first loads and inverts the input scribble image. The inverted image is then used by a "control_v11p_sd15_scribble_fp16" ControlNet model to extract and applied to the scribble, controlling and guiding the image generation process. A pre-trained model ("dreamshaper_8.safetensors") processes the positive conditioning generated by the combination of the text prompt and the ControlNet's output, along with negative conditioning from undesired traits such as "horror" and "lowres." The latent space image is generated using a KSampler and further decoded by the VAE to form the final image. The model synthesizes a highly detailed image while adhering closely to both the shape of the scribble and the desired features implied by the text prompt.
</principle>

## History

Here is a recent history of your thought, plan and action in the previous steps. The most recent record is at the bottom.

- Step: 1

<thought>
Since the history is empty, I need to start by loading a relevant reference workflow that can be used as a base for generating the 2-second video of colorful fish in an aquarium. The "text_to_video" reference seems appropriate to use as it involves generating video from text prompts, which aligns with the first part of the task.
</thought>

<plan>
Step 1: I will load the "text_to_video" reference workflow to use it as the base.  
Step 2: I will adapt the workflow to modify the prompt to match the requirements ("colorful fish swimming in an aquarium with bubbles rising").  
Step 3: I will combine it with the "video_frame_interpolation" reference to enable interpolation for smoother video.  
Step 4: I will adapt the interpolation multiplier to 3x to achieve the expected frame rate increase.  
Step 5: I will finish the task by confirming the generated output meets the required specifications.
</plan>

<action>
load(name="text_to_video")
</action>

- Step: 2

<thought>
The current workflow generates a 3-second video of a landscape scene, but this is not aligned with the task of generating a 2-second video of colorful fish swimming in an aquarium with bubbles rising. I need to adapt the text prompt accordingly to match the requirements. Additionally, the frame rate is currently 8 frames per second, but after generating the video, we will need to interpolate it to increase the frame rate by 3x to make the motion smoother.

\end{Verbatim}
\end{tcolorbox}

\clearpage

\begin{tcolorbox}[breakable, colback=bluebg, colbacktitle=bluebg, colframe=bluefg, fonttitle=\bfseries\color{black}, float*=ht, width=\textwidth]
\begin{Verbatim}[breaklines=true, breakanywhere=true, fontsize=\footnotesize]
The basic pipeline for generating videos from a text prompt is already functional, but the correct text needs to be provided, and the next step would be to apply video frame interpolation. I will combine this workflow with the "video_frame_interpolation" to achieve smoother motion.
</thought>

<plan>
Step 1: Adapt the text prompt to describe the required scene of "colorful fish swimming in an aquarium with bubbles rising".  
Step 2: Combine the current workflow with the "video_frame_interpolation" reference workflow to add video interpolation functionality.  
Step 3: Adapt the interpolation multiplier to 3x for smoother motion.  
Step 4: Finish the task after confirming that the workflow meets the requirements.
</plan>

<action>
adapt(prompt="Change the text prompt to 'colorful fish swimming in an aquarium with bubbles rising to the surface' and reduce the video duration to 2 seconds.")
</action>

- Step: 3

<thought>
The video generation part is complete, wherein a 2-second video of colorful fish swimming in an aquarium with bubbles rising is generated at 8 FPS. The next part of the task is to increase the frame rate by 3x using frame interpolation to achieve smoother motion. I need to combine the current workflow with the "video_frame_interpolation" reference to add the interpolation capability and then adjust the frame rate multiplier to 3x as per the requirements.
</thought>

<plan>
Step 1: Combine the current workflow with the "video_frame_interpolation" reference to add the interpolation functionality.  
Step 2: Adapt the multiplier for frame interpolation to 3x to ensure the frame rate increase.  
Step 3: Finish the task since this will meet the stated requirements.
</plan>

<action>
combine(name="video_frame_interpolation")
</action>

## Workspace

The code and annotation of the current workflow you are working on are presented as follows:

<code>
model_15, clip_vision_15, vae_15 = ImageOnlyCheckpointLoader(ckpt_name="""svd_xt_1_1.safetensors""")  
model_16, clip_16, vae_16 = CheckpointLoaderSimple(ckpt_name="""sd_xl_base_1.0.safetensors""")  
conditioning_18 = CLIPTextEncode(text="""colorful fish swimming in an aquarium with bubbles rising to the surface""", speak_and_recognation=True, clip=clip_16)  
conditioning_19 = CLIPTextEncode(text="""text, watermark""", speak_and_recognation=True, clip=clip_16)  
latent_22 = EmptyLatentImage(width=1024, height=576, batch_size=1)  
model_14 = VideoLinearCFGGuidance(min_cfg=1, model=model_15)  
latent_17 = KSampler(seed=307393744025667, steps=15, cfg=8, sampler_name="""uni_pc_bh2""", scheduler="""normal""", denoise=1, model=model_16, positive=conditioning_18, negative=conditioning_19, latent_image=latent_22)  
image_20 = VAEDecode(samples=latent_17, vae=vae_16)  
_ = PreviewImage(images=image_20)  
positive_12, negative_12, latent_12 = SVD_img2vid_Conditioning(width=1024, height=576, video_frames=16, motion_bucket_id=127, fps=8, augmentation_level=0, clip_vision=clip_vision_15, init_image=image_20, vae=vae_15)  
\end{Verbatim}
\end{tcolorbox}

\clearpage

\begin{tcolorbox}[breakable, colback=bluebg, colbacktitle=bluebg, colframe=bluefg, fonttitle=\bfseries\color{black}, float*=ht, width=\textwidth]
\begin{Verbatim}[breaklines=true, breakanywhere=true, fontsize=\footnotesize]
latent_3 = KSampler(seed=1103641334004632, steps=20, cfg=2.5, sampler_name="""euler""", scheduler="""karras""", denoise=1, model=model_14, positive=positive_12, negative=negative_12, latent_image=latent_12)  
image_8 = VAEDecode(samples=latent_3, vae=vae_15)  
filenames_23 = VHS_VideoCombine(frame_rate=8, loop_count=0, filename_prefix="""SVD_txt2vid""", format="""video/h264-mp4""", pix_fmt="""yuv420p""", crf=19, save_metadata=True, pingpong=False, save_output=True, images=image_8)

image_10 = RIFE_VFI(ckpt_name="""rife47.pth""", clear_cache_after_n_frames=10, ensemble=True, fast_mode=True, frames=image_8, multiplier=3, scale_factor=1)  
filenames_24 = VHS_VideoCombine(frame_rate=24, loop_count=0, filename_prefix="""SVD_txt2vid_Interpolated""", format="""video/h264-mp4""", pix_fmt="""yuv420p""", crf=19, save_metadata=True, pingpong=False, save_output=True, images=image_10)
</code>

<function>
This workflow first generates a 2-second video of colorful fish swimming in an aquarium with bubbles rising to the surface from a text description using a text-to-video approach. It then performs video interpolation to increase the frame rate by 3x, resulting in a smoother video.
</function>

<principle>
The workflow begins by loading the necessary models, including the Stable Diffusion model and the Stable Video Diffusion model. These are used to encode the input text to create a video showcasing colorful fish and bubbles. Initially, the video is generated at 8 frames per second (FPS) over 16 frames (~2 seconds). After the video is synthesized, it is passed through a frame interpolation process using the RIFE VFI model, which increases the frame rate by generating additional frames, multiplying the frame rate by 3x (from 8 FPS to 24 FPS). Finally, the interpolated frames are combined into a smoother MP4 video output.
</principle>

## Action

Based on the history and workspace, you should first think about what functions have been implemented and what modules remain to be added. Your thought should be enclosed with "<thought>" tag. For example: <thought>The basic pipeline has been implemented, but a module is needed to improve the quality.</thought>.

After that, you should update your step-by-step plan to further modify your workflow. There are 2 steps remaining, so your plan should contain at most 2 steps. Make sure that each step is feasible to be converted into a single action. Your plan should be enclosed with "<plan>" tag. For example: <plan>Step 1: I will refer to "reference_name" to add a module. Step 2: I will finish the task since the expected effects are realized.</plan>.

Finally, you should choose one of the following actions and specify the arguments (if required), so that the updated workflow can realize the first step in your plan. You should provide your action with the format of function calls in Python. Your action should be enclosed with "<action>" tag. For example: <action>combine(name="reference_name")</action>, <action>adapt(prompt="Change the factor to 0.5 and rewrite the prompt.")</action>, and <action>finish()</action>.

- `load`: Load a reference workflow into the workspace to replace the current workflow, so that you can start over. Arguments:
  - `name`: The name of the reference workflow you want to load.
- `combine`: Combine the current workflow with a reference workflow, so that necessary modules can be added. Arguments:
  - `name`: The name of the reference workflow you want to combine.
- `adapt`: Adapt some parameters in the current workflow, so that the expected effects can be realized. Arguments:
  - `prompt`: The prompt to specify the adaptation you want to make.
- `retrieve`: Retrieve a new batch of reference workflows, so that more useful references can be found. Arguments:
  - `prompt`: The prompt to describe the reference you want to retrieve.
- `finish`: Finish the task since the current workflow can realize the expected effects.

Refer to the history before making a decision. Here are some general rules you should follow:

1. You should choose the `load` action if and only if the history is empty.
2. If you choose the `load` or `combine` action, make sure the name exists in the reference. Otherwise, try to update the reference with the `retrieve` action.
3. You should not choose the `adapt` action twice in a row, because they can be simplified into a single action.
4. If you choose the `adapt` or `retrieve` action, make sure the prompt is concise and contains all the necessary information.
5. You should choose the `finish` action before the remaining steps count down to 0.

Now, provide your thought, plan and action with the required format.
\end{Verbatim}
\end{tcolorbox}

\begin{tcolorbox}[title=PlanAgent Answer, breakable, colback=redbg, colbacktitle=redbg, colframe=redfg, fonttitle=\bfseries\color{black}, float*=ht, width=\textwidth]
\begin{Verbatim}[breaklines=true, breakanywhere=true, fontsize=\footnotesize]
<thought>The current workflow successfully generates a 2-second video of colorful fish swimming in an aquarium with bubbles rising to the surface and then applies video interpolation to enhance the frame rate by 3x. The workflow seems to fulfill both the generation and interpolation requirements. No additional modifications appear to be necessary, as the video generation part is functional and the interpolation process has been properly set to increase the frame rate from 8 FPS to 24 FPS.</thought>

<plan>  
Step 1: I will finish the task since the expected effects have been realized.
</plan>

<action>finish()</action>
\end{Verbatim}
\captionsetup[figure]{hypcap=false}
\captionof{figure}{A typical example trajectory of ComfyAgent.}
\label{figure:example_trajectory}
\end{tcolorbox}

\end{document}